\documentclass[sigconf]{acmart}
\usepackage{booktabs} 
\usepackage[font=small]{caption}
\usepackage[labelformat=simple]{subcaption}
\usepackage{graphicx}
\usepackage[ruled,vlined,linesnumbered]{algorithm2e}
\usepackage{algorithmic}
\SetKwComment{Comment}{$\triangleright$\ }{}
\usepackage{indentfirst}
\usepackage{mdwlist}
\usepackage[mathscr]{eucal}
\usepackage{changepage}
\usepackage{color}
\usepackage{url}
\usepackage{hyphenat}
\usepackage[font=small]{caption}
\usepackage{makecell}
\usepackage{mathtools}
\usepackage{xcolor}
\usepackage{cleveref}
\usepackage{footnote}
\usepackage{multirow}
\usepackage{soul}
\usepackage{times}
\usepackage{etoolbox}
\usepackage{textcomp}

\renewcommand{\textrightarrow}{$\rightarrow$}

\allowdisplaybreaks

\newcommand{\method}{\textsc{DAIN}\xspace}

\newcommand{\tracin}{\textsc{TracIn}\xspace}
\newcommand{\costco}{\textsc{CoSTCo}\xspace}

\newcommand{\T}[1]{\boldsymbol{\mathscr{#1}}}   
\newcommand{\tensor}[1]{\boldsymbol{\mathscr{#1}}}   
\newcommand{\mat}[1]{\mathbf{#1}}

\newcommand{\argmin}[1]{\underset{#1}{\operatorname{arg}\,\operatorname{min}}\;}

\copyrightyear{2021} 
\acmYear{2021} 
\setcopyright{acmcopyright}
\acmConference[CIKM '21]{Proceedings of the 30th ACM International Conference on Information and Knowledge Management}{November 1--5, 2021}{Virtual Event, QLD, Australia}
\acmBooktitle{Proceedings of the 30th ACM International Conference on Information and Knowledge Management (CIKM '21), November 1--5, 2021, Virtual Event, QLD, Australia}
\acmPrice{15.00}
\acmDOI{10.1145/3459637.3482267}
\acmISBN{978-1-4503-8446-9/21/11}

\begin{document}
\fancyhead{}

\title{Influence-guided Data Augmentation for\\ Neural Tensor Completion}

\author{Sejoon Oh}
\email{soh337@gatech.edu} 
\affiliation{%
  \institution{Georgia Institute of Technology}
    \country{United States}
}

\author{Sungchul Kim, Ryan A. Rossi}
\email{sukim@adobe.com, ryrossi@adobe.com} 
\affiliation{%
  \institution{Adobe Research}
    \country{United States}
}

\author{Srijan Kumar}
\email{srijan@gatech.edu} 
\affiliation{%
  \institution{Georgia Institute of Technology}
    \country{United States}
}

\renewcommand{\shortauthors}{Oh, et al.}

	\begin{abstract}
		\label{sec:abstract}
		\noindent
How can we predict missing values in multi-dimensional data (or tensors) more accurately?
The task of tensor completion is crucial in many applications such as personalized recommendation, image and video restoration, and link prediction in social networks.
Many tensor factorization and neural network-based tensor completion algorithms have been developed to predict missing entries in partially observed tensors. 
However, they can produce inaccurate estimations as real-world tensors are very sparse, and these methods tend to overfit on the small amount of data.
Here, we overcome these shortcomings by presenting a data augmentation technique for tensors.
In this paper, we propose \method, a general data augmentation framework that enhances the prediction accuracy of neural tensor completion methods.
Specifically, \method first trains a neural model and finds tensor cell importances with influence functions.
After that, \method aggregates the cell importance to calculate the importance of each entity (\textit{i.e.}, an index of a dimension). Finally, \method augments the tensor by weighted sampling of entity importances and a value predictor. 
Extensive experimental results show that \method outperforms all data augmentation baselines in terms of enhancing imputation accuracy of neural tensor completion on four diverse real-world tensors. Ablation studies of \method substantiate the effectiveness of each component of \method.
Furthermore, we show that \method\ scales near linearly to large datasets. 
\vspace{-2mm}
	\end{abstract}



 \maketitle
	
	\section{Introduction}
	\label{sec:intro}
	We observe various tensors on the Web, including images and videos, numerical ratings, social networks, and knowledge bases.
Most real-world tensors are very large and extremely sparse~\cite{ptucker} (\textit{i.e.}, a few observed entries and many missing values). Thus, tensor completion, \textit{i.e.}, the task of predicting missing values in a tensor, has been actively investigated in diverse areas including recommender systems~\cite{frolov2017tensor}, social networks~\cite{dunlavy2011temporal}, traffic analysis~\cite{tan2013tensor}, medical questionnaires~\cite{dauwels2012tensor}, and computer vision ~\cite{dian2017hyperspectral}.
For example, consider a movie rating tensor with three dimensions, namely users, movies, and time slices. Each tensor cell $(i,j,k)$ contains the rating score given by user $i$ to movie $j$ during time slice $k$. 
The goal of tensor completion is to predict the rating scores of unobserved tensor cells. 

Tensor factorization (TF) is a popular technique to predict missing values in a tensor, but most methods~\cite{MET,kaya2015scalable,smith2017tucker,  oh2017s, ShinK17, ptucker} exhibit high imputation error while estimating missing values in a tensor. One of the reasons is that many TF models~\cite{MET, kaya2015scalable, oh2017s, smith2017tucker} regard missing values in a tensor as zeros. Hence, if those models are trained with a sparse tensor, their predictions would be biased toward zeros, instead of the observed values.
Other TF methods~\cite{filipovic2015tucker, ShinK17,ptucker} improve their accuracy by focusing only on observed entries; however, they suffer from overfitting as the tensor is very sparse. 

Neural network-based tensor completion methods~\cite{NTF2019, costco2019, NTM, xie2020neural} have been proposed to enhance the estimation accuracy. They have strong generalization capability~\cite{neyshabur2017exploring}, and capture non-linearity hidden in a tensor. However, these methods still suffer from data sparsity,  and this can become a bottleneck for neural tensor completion methods, which require a large amount of data for training~\cite{wong2016understanding}. Moreover, these existing methods cannot generate new data points for data augmentation to solve the sparsity issue. 

In this paper, we propose a method that leverages the strength of neural tensor completion, and improves it through the utilization of data augmentation. 
Data augmentation increases the generalization capability of a model by generating new data points while training and has shown success in various applications of deep learning~\cite{Shorten2019, perez2017effectiveness, Xu2016ImprovedRC}; however, it has not been explored in the task of tensor completion.
Here, we propose an influence-guided data augmentation technique called \method (\underline{D}ata \underline{A}ugmentation with \underline{IN}fluence Functions). 
First, \method trains a neural tensor completion model with the input tensor, and utilizes influence functions to estimate the importance of each training cell (\textit{e.g.}, the importance of a rating in a movie rating tensor) on reducing imputation error. 
Next, \method\ computes the importance of every entity by aggregating the importance values of all its associated cells.
For example, to compute the importance of a user $i$, the importance of all the ratings given by $i$ are aggregated. 
The importance of an entity signifies its impact in reducing the prediction error.
Finally, \method generates new data points by sampling entities proportional to their importance scores. Values of the augmented tensor cells are  predicted via a trained neural tensor completion method. 
This influence-based sampling of entities augments data points using important entities, and thus, can lead to higher test prediction accuracy.

We evaluate \method\ by conducting extensive experiments on four diverse real-world datasets and one synthetic dataset. 
Results show that \method\ outperforms baseline augmentation methods with statistical significance, and that \method\ improves the prediction performance of a neural tensor completion method as the number of augmentation increases.
Via thorough ablation studies, we confirm the effectiveness of each component of \method.
We also show that \method\ is scalable to large datasets as illustrated by a near-linear relationship between runtime and the amount of augmentation.
Finally, we demonstrate the sensitivity of the performance of \method\ with respect to hyperparameter values.

Our main contributions are summarized by the following:
\begin{itemize}
\item \textbf{Novel problem.}
We propose a novel data augmentation technique \method for enhancing neural tensor completion. To the best of our knowledge, this is the first data augmentation method for tensor completion.
\item \textbf{Algorithmic framework.} 
We propose a new framework for deriving the importance of tensor entities on reducing prediction error using influence functions. With the entity importance values, we create new data points via weighted sampling and value predictions. We also provide complexity analyses of \method.
\item \textbf{Performance.} 
\method outperforms baseline data augmentation methods on various real-world tensors in terms of prediction accuracy with statistical significance. 
\end{itemize}

The code and datasets are available on the project website:\\ \texttt{https://github.com/srijankr/DAIN}.
	
	\section{Related Work}
	\label{sec:related_work}
	
{
\small
\begin{table}[t]
	\centering
	\caption{Comparison of our proposed data augmentation framework \method against existing methods. A checkmark indicates that a method satisfies a specific criterion. \method is the only data augmentation method satisfying all criteria.}
	\begin{tabular}{c|ccccc|c}
		\toprule
		&  \rotatebox[origin=l]{90}{\parbox{2.0cm}{NTF~\cite{NTF2019}}} &  \rotatebox[origin=l]{90}{\parbox{2.0cm}{\costco~\cite{costco2019}}} &  \rotatebox[origin=l]{90}{\parbox{2.0cm}{NTM~\cite{NTM} }}  &
		\rotatebox[origin=l]{90}{\parbox{2.0cm}{\tracin~\cite{tracin}}}  &
		\rotatebox[origin=l]{90}{\parbox{2.0cm}{Aug-Net~\cite{Lee_2020_CVPR}}} 
		& \rotatebox[origin=l]{90}{\parbox{2.0cm}{\textbf{\method \\ (Proposed)}}} 
		 \\
		\midrule
        Neural Tensor Completion & \checkmark & \checkmark  & \checkmark  & &  & \checkmark \\
		Architecture Generalizability & &   &   &   \checkmark &  \checkmark & \checkmark \\
		Influence Utilization & &   &   &  \checkmark & \checkmark & \checkmark \\
		Data Augmentation Ability &  &   &   &   &  \checkmark & \checkmark  \\
		\bottomrule
	\end{tabular}
	\label{tab:comparators}
\end{table}
}

\textbf{Neural Tensor Completion.}
After recent advances in neural networks (NN), there have been many research topics applying NN to tensor completion frameworks. 
Neural tensor factorization (NTF)~\cite{NTF2019} is the first model that employs MLP and long short-term memory (LSTM) architectures for a tensor completion task. 
Neural Tensor Machine (NTM)~\cite{NTM} combines the outputs from generalized CANDECOMP/PARAFAC (CP) factorization and a tensorized MLP model to estimate missing values.
Moreover, Liu \textit{et al.}~\cite{costco2019} propose a convolutional neural network (CNN)-based tensor completion model, named \costco, which learns entity embeddings via several convolutional layers. 
However, as shown in Table~\ref{tab:comparators}, the above methods cannot augment new data points, do not generalize to multiple neural network architectures, and do not utilize influence functions. \method can provide high-quality augmentation for these methods while having these desirable properties. 
Note that we exclude tensor completion methods~\cite{ptucker,liu2014generalized,yang2021mtc,ShinK17} that do not utilize neural networks since \method is designed for the neural methods.

\textbf{Influence Estimation.} 
Computing the influence (or importance) of data points on model predictions or decisions has been actively investigated~\cite{KohL17, yeh2018representer,ghorbani2019data, tracin}. Influence has been mainly used to explain the prediction results by identifying useful or harmful data points with respect to the predictions. Firstly, Yeh \textit{et al.}~\cite{yeh2018representer} leverage the Representer Theorem for explaining deep neural network predictions. Koh and Liang~\cite{KohL17} introduce a model-agnostic approach based on influence functions. Amirata and James~\cite{ghorbani2019data} propose a data Shapley metric to quantify the value of each training datum in the context of supervised learning.  
More recently, \tracin~\cite{tracin} is proposed to compute the influence of a training example on a prediction made by a model with training and test loss gradients.
In this paper, we decide to use \tracin for the tensor data augmentation setting since it is scalable, easy to implement,
and does not rely on optimality conditions as Influence Functions~\cite{KohL17} or Representer Theorem~\cite{yeh2018representer} does.

\textbf{Data Augmentation.}
Data augmentation is widely used to significantly increase the data available for training models without collecting new data~\cite{ho2019population}.
Basic image data augmentation methods include geometric transformations (such as flipping, cropping, rotation, and translation), mixing images by averaging their pixel values~\cite{inoue2018data}, and random erasing which randomly selects an $n \times m$ patch of an image and masks it with certain values~\cite{zhong2020random}. 
Augmentation Network model~\cite{Lee_2020_CVPR} finds image augmentation that maximizes the generalization
performance of the classification model via trainable image transformation models and influence functions. 
Earlier data augmentation methods for text data are based on synonym replacement - replacing a random word in a given sentence with its synonym by using external resources such as WordNet~\cite{zhang2015character, mueller2016siamese} and the pre-trained language model~\cite{jiao2020tinybert}. Yuning \textit{et al.}~\cite{xie2019unsupervised} leverage machine translation to paraphrase a given text. 
There is a line of research that perturb text via regex-based transformation~\cite{coulombe2018text}, noise injection~\cite{XieWLLNJN17}, and combining words~\cite{guo2019augmenting}.
However, none of the above methods are applicable in a tensor completion task.
To the best of our knowledge, ours is the first work that enhances the performance of neural tensor completion via data augmentation.

	\section{Preliminaries}
	\label{sec:preliminary}
	In this section, we define the preliminary concepts necessary to understand the proposed method. Table~\ref{tab:Symbols} summarizes the symbols frequently used in the paper.

\subsection{Tensor}
\label{sec:prelim:tensor}
Tensors are multi-dimensional data and a generalization of vectors ($1$-order tensors) and matrices ($2$-order tensors) to the higher order.
An $N$-way or $N$-order tensor has $N$ dimensions, and the dimension size (or dimensionality) is denoted by $I_{1}$ through $I_{N}$, respectively.
We denote an $N$-order tensor by boldface Euler script letters (\textit{e.g.}, $\tensor{X} \in \mathbb{R}^{I_{1} \times \cdots \times I_{N}}$).
A tensor cell $(i_{1},...,i_{N})$ contains the value $\tensor{X}_{(i_{1},...,i_{N})}$, and an entity of a tensor refers to a single index of a dimension. 
For example, in the movie rating tensor case, an entity refers to a user, movie, or time slice, and a tensor cell contains a rating. 

\begin{table}[t!]
	\small
	\centering
	\caption{Table of symbols.}
	\hspace{-5mm}
	\begin{tabular}{c | l}
		\toprule
		\textbf{Symbol} & \textbf{Definition} \\
		\midrule
		$\tensor{X}$ & input tensor\\
		$N$ & order of $\tensor{X}$\\
		$I_{n}$ & dimensionality/size of the $n^{th}$ dimension of
		$\tensor{X}$\\
        $(i_1,...,i_N)$ &  cell of $\tensor{X}$ \\
		$i_n$ &  entity of the $n^{th}$ dimension of $\tensor{X}$ \\
		$\Omega_{train}$ & set of train cells of $\tensor{X}$\\
		$\Omega_{val}$ & set of validation cells of $\tensor{X}$\\
		$\Omega_{test}$ & set of test cells of $\tensor{X}$\\
		$\tensor{\alpha}$ & cell importance tensor \\
		$\alpha^{(1)}, ... , \alpha^{(N)}$ & entity importance \\
        $E_{i}^{n}$ & embedding of an entity $i$ of the $n^{th}$ dimension \\
		$\Theta$ & parameters of a tensor completion model\\
		$\Theta_t$ & parameters of a tensor completion model at epoch $t$\\
		$\eta_i$ & step size at a checkpoint $\Theta_{t_i}$\\
		$\Theta_{t_1}, ..., \Theta_{t_K}$ & $K$ checkpoints saved at epochs $t_1, ... , t_K$ \\
        $N_{aug}$ & number of data augmentation \\
        $T_{\Theta}, M_{\Theta}$ & time and space complexity of training entity embeddings \\
        $T_{\Theta_{p}}, M_{\Theta_{p}}$ & time and space complexity of training a value predictor \\
        $T_{infer}$ & time complexity of a single inference of a value predictor	 \\
        $D$ & dimension of a gradient vector \\	
		\bottomrule
	\end{tabular}
	\label{tab:Symbols}
\end{table}

\subsection{Tensor Completion}
\label{sec:prelim:ntf}
Tensor completion is defined as the problem of filling the missing values of a partially observed tensor. Tensor completion methods train their model parameters with observed cells ($\Omega_{train}$) and predict values of unobserved cells ($\Omega_{test}$) with the trained parameters.
Specifically, given an N-order tensor $\tensor{X}~(\in \mathbb{R}^{I_{1} \times \cdots \times I_{N}})$ with training data $\Omega_{train}$,  a tensor completion method aims to find model parameters $\Theta$ for the following optimization problem.
\begin{equation}
    \label{eq:nn_objective}
    \argmin {\Theta} {\sum_{\forall (i_1, ..., i_N) \in \Omega_{train}} \left(\T{X}_{(i_1, ... i_N)}-\hat{\T{X}}_{(i_1,...,i_N)}\right)^2}
\end{equation}
where $\hat{\T{X}}_{(i_1,...,i_N)} = \Theta(i_1,...,i_N)$ is a prediction value for a cell $(i_1,...,i_N)$ generated by the tensor completion method $\Theta$. 
Neural tensor completion methods utilize different neural network architectures to compute $\hat{\T{X}}_{(i_1,...,i_N)}$ (see Section~\ref{sec:NN_training} for the multilayer perceptron case). 
Root-mean-square error (RMSE) is a popular metric to measure the accuracy of a tensor completion method~\cite{costco2019, NTF2019, ptucker}. Specifically, we use test RMSE to check how accurately a tensor completion model predicts values of unobserved tensor cells. The formal definition of test RMSE is given as follows. 
Notice that a tensor completion model with the lower test RMSE is more accurate.
    \begin{equation} \label{eq:rmse}
        Test-RMSE = \sqrt{\frac{1}{|\Omega_{test}|}\sum_{\forall(i_1,...,i_N)\in\Omega_{test}}{\left(\tensor{X}_{(i_1,...,i_N)}-\hat{\T{X}}_{(i_1,...,i_N)}\right)}^2}
    \end{equation}

\subsection{Influence Estimation with \tracin}
\label{sec:prelim:tracin}
We introduce the state-of-the-art influence estimator: \tracin~\cite{tracin}.
\tracin calculates the importance of every training data point in reducing test loss. This is done by tracing training and test loss gradients with respect to model checkpoints, where checkpoints are the model parameters obtained at regular intervals during the training (\textit{e.g.}, at the end of every epoch). 
The influence of a training data point $z$ on the loss of a test data point $z'$ is given as follows (please refer to Section 3 of \tracin~\cite{tracin} for the details):
\begin{equation}
\label{eq:influence_test}
Inf(z,z') \approx \sum_{i=1}^{K} \eta_{i} \nabla \ell(\Theta_{t_i},z) \cdot \nabla \ell(\Theta_{t_i},z')  
\end{equation}
where $\Theta_{t_i}, 1 \le i \le K$ are checkpoints saved at epochs $t_1, ..., t_K$, $\eta_i$ is a step size at a checkpoint $\Theta_{t_i}$, and $\nabla \ell(\Theta_{t_i},z)$ is the gradient of the loss of $z$ with respect to a checkpoint $\Theta_{t_i}$.

Influence estimation with \tracin has been shown to have clear advantages in terms of speed and accuracy over existing methods~\cite{tracin}, such as Influence Functions~\cite{KohL17} or Representer Point method~\cite{yeh2018representer}. 
We utilize \tracin to create the Cell Importance Tensor in Section~\ref{sec:instance_importance}.

	\section{Proposed Method: \method}
	\label{sec:proposed_method}
	
\begin{figure*}[t!]
	\centering
	\includegraphics[width=0.76\textwidth]{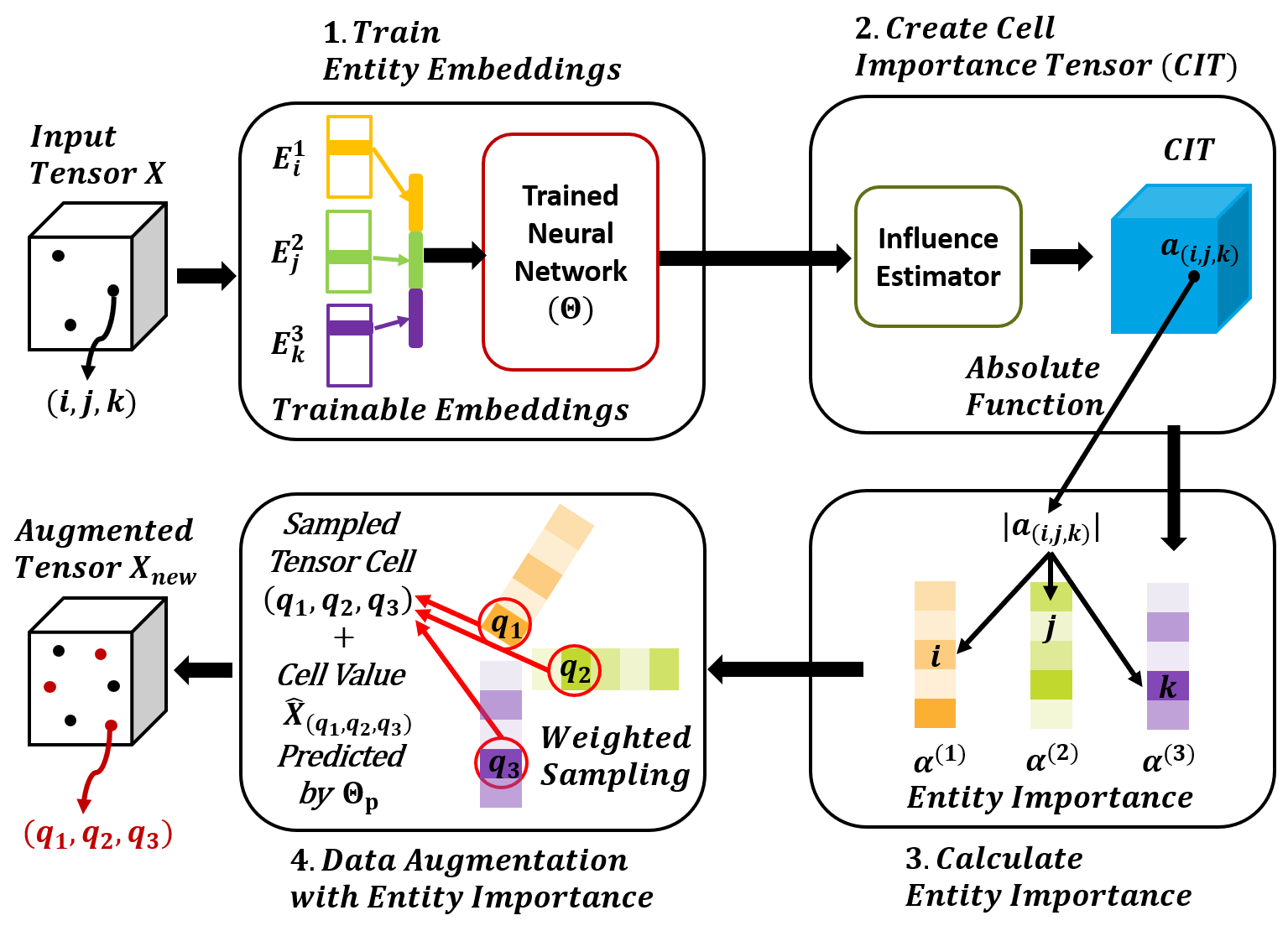}
	\caption{
	An overview of our proposed data augmentation method \method for a 3-dimensional tensor. \method first learns entity embeddings with training data and computes training and validation loss gradients (step 1). 
	Next, it obtains cell importance values by the influence estimator \tracin and creates the Cell Importance Tensor (step 2). 
	\method uniformly distributes cell importance values to the corresponding entities, and each entity calculates its importance by aggregating the assigned cell importance values (step 3).
	Finally, \method performs data augmentation to an input tensor by weighted sampling on those entity importance arrays and value predictions (step 4). Note that \method can use any neural tensor completion methods to learn entity embeddings (step 1) or predict values of sampled tensor cells (step 4).
	Please refer to each subsection in Section~\ref{sec:proposed_method} for detailed information on each step.
	}
	\label{fig:mainplot}
\end{figure*}

In this section, we describe \method, a data augmentation pipeline using the importance of entities for efficient tensor completion. 
Figure~\ref{fig:mainplot} shows our proposed method, and Algorithm~\ref{alg:main} summarizes the overall data augmentation process of \method. 
We explain the steps of our method in the next subsections.

\subsection{Training Entity Embeddings}
\label{sec:NN_training}
First, we train a neural network model to learn embeddings for every entity in an input tensor (\textit{e.g.}, user, movie, and time embeddings in the movie rating tensor).
The traditional technique of learning one embedding per data point is not scalable in tensors, as the number of data points in the tensor can be very large.
Instead, we learn one embedding per entity of each dimension (\textit{i.e.}, in the movie rating tensor example, we learn one embedding for every user, movie, and time slice, respectively).
Each tensor cell can then be represented as an ordered concatenation of the embeddings of its entities. 
For example, a tensor cell $(i_1, ... , i_N)$ can be represented as [$E_{i_1}^{1}$, $E_{i_2}^{2}$, $E_{i_3}^{3}, \ldots$], where $E_{i_n}^{n}$ is an embedding of an entity $i_n$ of the $n^{th}$ dimension. 
The entity embeddings are trained to accurately predict the values of the (training) data points in the tensor. 

We train an end-to-end trainable neural network to learn the entity embeddings (line 1 in Algorithm~\ref{alg:main}). 
We choose a multilayer perceptron (MLP) model with ReLU activation since it shows the best imputation performance with data augmentation (see  Figures~\ref{fig:ablation_embedding_mnist} and \ref{fig:ablation_embedding_foursquare}).
Note that existing neural tensor completion models, such as NTF~\cite{NTF2019} and CoSTCo~\cite{costco2019}, can be also used to generate these embeddings. 
Our model's prediction value for a tensor cell $(i_1, ... , i_N)$ is defined by the following.
\begin{equation}
\label{eq:mlp}
		\begin{aligned}
	& Z_1 = \phi_1 (W_1 [E_{i_1}^{1},\ldots,E_{i_N}^{N}]+b_1), \cdots ,\\
	& Z_{M} = \phi_{M} (W_{M} Z_{M-1}+b_{M}) \\
	& \hat{\T{X}}_{(i_1,...,i_N)} = W_{M+1} Z_{M} + b_{M+1}
		\end{aligned}
	\end{equation}

\noindent Note that $[E_{i_1}^{1},\ldots,E_{i_N}^{N}]$ represents the embedding of a cell $(i_1, ... , i_N)$ obtained by concatenating embeddings of entities $i_1, ... , i_N$, $\hat{\T{X}}_{(i_1,...,i_N)}$ is the imputed output value by the neural network for a cell $(i_1, ... , i_N)$, and $M$ indicates the number of hidden layers.  $W_1, ..., W_{M+1}$ and $b_1, ... , b_{M+1}$ are weight matrices and bias vectors, respectively. $\phi_1, ..., \phi_{M}$ are activation functions (\method uses $ReLU$). The model parameters $W_1, ..., W_{M+1}$ and $b_1, ... , b_{M+1}$ are referred to as $\Theta$.
By minimizing the loss function~\eqref{eq:nn_objective} combined with Equation~\eqref{eq:mlp}, we obtain trained entity embeddings as well as loss gradients for all training and validation cells (lines 2-3 in Algorithm~\ref{alg:main}). We utilize those gradient vectors to compute cell-level importance values in the next part.

\subsection{Creating Cell Importance Tensor}
\label{sec:instance_importance}
In Step 2, we create Cell Importance Tensor (CIT) which represents the importance of every tensor cell in reducing the prediction loss. In the movie rating tensor example, CIT stores the importance of ratings.
Any influence estimator can be used to compute the CIT, but we choose \tracin~\cite{tracin} introduced in Section~\ref{sec:prelim:tracin} since it is scalable, easy to implement, and does not rely on optimality conditions. Moreover, we later show \method with \tracin shows the highest prediction performance empirically compared to other influence estimation methods (see Figures~\ref{fig:compare_inf_mnist} and \ref{fig:compare_inf_foursquare}). 
We derive a value of a CIT cell $z$  by the following steps. Computing the CIT is shown in step 2 in Figure~\ref{fig:mainplot}.

Equation~\eqref{eq:influence_test} in Section~\ref{sec:prelim:tracin} computes the influence of a training cell $z$ on the loss of a test cell $z'$. Since we cannot access the test data, we compute the influence $\tensor{\alpha}_{z}$ of a training cell $z$ on reducing overall validation loss by the following (line 4 in Algorithm~\ref{alg:main}). 
\begin{equation}
    \label{eq:instance:importance}
    \begin{aligned}
    \tensor{\alpha}_{z} = |\sum_{z' \in \Omega_{val}} Inf(z,z')| = |\sum_{z' \in \Omega_{val}} \sum_{i=1}^{K} \eta_{i} \nabla \ell(\Theta_{t_i},z) \cdot \nabla \ell(\Theta_{t_i},z')| \\
 \Longleftrightarrow \tensor{\alpha}_{z} = |\sum_{i=1}^{K} \eta_{i} \nabla \ell(\Theta_{t_i},z) \cdot \bigg(\sum_{z' \in \Omega_{val}} \nabla \ell(\Theta_{t_i},z')\bigg)  | 
    \end{aligned}
\end{equation} 
where $K$ is the number of checkpoints, $\eta_{i}$ is a step size at a checkpoint $\Theta_{t_i}$, and $\nabla \ell(\Theta_{t_i},z)$ is the gradient of the loss of $z$ with respect to a checkpoint $\Theta_{t_i}$.
We use an absolute value function to the $\tensor{\alpha}_{z}$ calculation since cells with negative influence can also be important. For example, cells with large negative influence contribute to increasing the validation loss significantly. We can mitigate the loss increase by our data augmentation method, where the absolute function leads \method to create more augmentation for these cells.

\begin{algorithm} [t!]
	\footnotesize
	\caption{Data Augmentation with \method} \label{alg:main}
	\SetKwInOut{Input}{Input}
	\SetKwInOut{Output}{Output}
	\Input{
		Tensor $\T{X} \in \mathbb{R}^{I_1 \times I_2 \times \cdots \times I_N}$, 
		training tensor cells $\Omega_{train}$, validation tensor cells $\Omega_{val}$, 
		entity embedding training model $\Theta$,
		tensor value prediction model $\Theta_{p}$, 
		training checkpoints $\Theta_{t_1}, ..., \Theta_{t_K}$, step size $\eta_i$, and
		the maximum number of augmentation $N_{aug}$. \\
	}
	\Output{
		Augmented tensor $\T{X}_{new} = \T{X} \cup \T{X}_{aug}$.
	}
	\Comment{\textbf{Section~\ref{sec:NN_training}}}{
	Train $\Theta$ by Equation~\eqref{eq:nn_objective}   \\
	Obtain entity embeddings for all entities\\ 
	Obtain training and validation loss gradients for all training checkpoints   \\
	}
	\Comment{\textbf{Section~\ref{sec:instance_importance}}}{
	Create Cell Importance Tensor $\tensor{\alpha}$ by Equation~\eqref{eq:instance:importance} with $\Omega_{train}$ and $\Omega_{val}$\\
	}
	\Comment{\textbf{Section~\ref{sec:method:entity_importance}}}{
	Compute the entity importance $\alpha_{i}^{(n)}$ of all entities by Equation~\eqref{eq:entity_importance} \\
	}
	\Comment{\textbf{Section~\ref{sec:data_augmentation}}}{
	Train $\Theta_{p}$ by Equation~\eqref{eq:nn_objective}\\
	$\Omega_{aug} = \emptyset$ \\
	\For{$i = 1, \ldots, N_{aug}$}{
	    $z_{new} = \emptyset$ \\
	    \For{$n = 1, \ldots, N$}{
	        $i_n \longleftarrow$ weighted sampling on $\alpha^{(n)}$ (probability:  $\frac{\alpha_{i_n}^{(n)}}{\sum_{i=1}^{I_n} \alpha_{i}^{(n)}}$)\\
	        $z_{new} = z_{new} \cup \{i_n\}$
	    }
	    $\Omega_{aug} = \Omega_{aug} \cup \{z_{new}\}$
	}
	let $\T{X}_{aug}$ be a tensor in $\mathbb{R}^{I_1 \times I_2 \times \cdots \times I_N}$ with cells $\Omega_{aug}$, where the values of $\Omega_{aug}$ are imputed by the trained value predictor $\Theta_{p}$ \\
	$\T{X}_{new} = \T{X} \cup \T{X}_{aug}$ \\
	}
\end{algorithm}

\subsection{Calculating Entity Importance}
\label{sec:method:entity_importance}
Identifying important entities is crucial.
The output of the previous step identifies the cell importance, but the importance of each entity remains unknown.
For instance, in the movie rating tensor example, the cell importance score captures the importance of the rating on the prediction loss, while it does not reflect the importance of a user, a movie, or a time slice.
If we can find the importance of every user, movie, and time slice, we can generate new influential data points to minimize prediction error by combining users, movies, and time slices that have high importance values.

Here, we describe an aggregation technique to calculate the entity importance values from cell importance values (shown in  step 3 of Figure~\ref{fig:mainplot} and line 5 in Algorithm 1). First, we uniformly distribute a cell's importance value to its associated entities. For instance, given a training cell $z=(i_1, ..., i_N)$, we uniformly distribute $\tensor{\alpha}_{z}$ to its $N$ entities $\{i_1, ... , i_N\}$. After performing the allocation for all training cells,  we compute the entity importance score $\alpha_{i}^{(n)}$ for an entity $i$ of the $n^{th}$ dimension by aggregating cell importance scores as follows:
\begin{equation}
\alpha_{i}^{(n)} = \sum_{\forall (i_1, ..., i_N) \in \Omega_{train}, i_{n}=i} \tensor{\alpha}_{(i_1, ..., i_N)}
\label{eq:entity_importance}
\end{equation}
Recall that $\tensor{\alpha}$ indicates the Cell Importance Tensor, and $\Omega_{train}$ represents a set of training cells from an original tensor. In the movie rating tensor example, the above equation signifies that a user's entity importance is the aggregation of the importance scores of all the ratings the user gives. Similarly, a movie's importance is the sum of the importance of the ratings it receives, and the importance of a time slice is the sum of the importance of all the ratings given during the time slice. 

Another way of computing the entity importance is applying rank-1 CP (CANDECOMP/PARAFAC) factorization~\cite{kolda2009tensor} on the Cell Importance Tensor. The output factor matrices from the CP model are entity importances.
Specifically, a value of each output array indicates the importance of the corresponding entity on predicting values in a training tensor. 
The loss function of the rank-1 CP model is given as follows.
	\begin{multline}
		L(\mat{\alpha}^{(1)},...,\mat{\alpha}^{(N)}) = \\
		\hspace{-4mm}
		 \sum_{\forall(i_1,...,i_N)\in\Omega_{train}}{\left(|\tensor{\alpha}_{(i_1,...,i_N)}|-\prod_{n=1}^{N} \alpha^{(n)}_{i_{n}}\right)^{2}} +  \lambda \sum_{n=1}^{N}{{\| \mat{\alpha}^{(n)} \|}^{2}}
		\label{eq:alpha_factorization}
		\end{multline}
Note that $\alpha^{(1)},...,\alpha^{(N)}$ are entity importances,  $\lambda$ is a regularization factor, and $\|\tensor{X}\|$ is Frobenius norm of a tensor $\T{X}$.
We choose the aggregation scheme to compute the entity importance as the rank-1 CP model can produce inaccurate decomposition results when a given tensor is highly sparse. By conducting extensive experiments, we later show that the aggregation method over the rank-1 CP model with respect to the prediction accuracy (see Figures~\ref{fig:compare_entity_mnist} and \ref{fig:compare_entity_foursquare}).

\subsection{Data Augmentation with Entity Importance}
\label{sec:data_augmentation}
In the final step, we identify the tensor cells and values for data augmentation using the entity importance scores and a value predictor, respectively. This is illustrated in step 4 of Figure~\ref{fig:mainplot} and lines 6--15 in Algorithm 1. A high entity importance score signifies that the corresponding entity plays an important role in improving the validation set prediction. Thus, we create new cells using these important entities.

We first train a neural tensor completion model $\Theta_{p}$ with the original training cells (line 6), which will be used later for value predictions. 
After that, we conduct weighted sampling on every entity importance array (line 11) and select one entity from each dimension (line 12). Mathematically, an entity $i$ of the $n^{th}$ dimension (\textit{e.g.}, a user among all users) has a probability $\frac{\alpha_{i}^{(n)}}{\sum_{i=1}^{I_n} \alpha_{i}^{(n)}}$ to be sampled.
We combine the sampled entities from all dimensions to form one tensor cell (line 13).
We repeat the process several times to create the required number of data points for augmentation (lines 8--13).

Once indices of the cells are sampled, their values need to be determined (line 14). Trivially, one can use the overall average value (\textit{i.e.}, $\frac{1}{|\Omega_{train}|}\sum_{\forall(i_1,...,i_N)\in\Omega_{train}}{\tensor{X}_{(i_1,...,i_N)}}$) or find the most similar index in the embedding space and take its value. However, these heuristics can be inaccurate and computationally expensive, respectively. 
An advanced way of assigning the values of the augmented data points is by predicting the values using a tensor completion model (either the previously trained $\Theta$ or training another model $\Theta_{p}$). Specifically, we can use the trained embeddings from $\Theta$ to predict the values, or we can train an off-the-shelf prediction method $\Theta_{p}$ with the training data and predict the values with it.  
Reusing the trained neural network $\Theta$ is computationally cheaper since it only needs to do a forward pass for inference, which is fast. However, this can cause overfitting in the downstream model since the resulting augmentation cell values are likely to be homogeneous to the original tensor. 
On the other hand, using another off-the-shelf method $\Theta_{p}$ can increase the generalization capability of a downstream model by generating more heterogeneous data compared to  $\Theta$. 
Thus, we choose to train a new off-the-shelf model $\Theta_{p}$.
We use the state-of-the-art algorithm \costco~\cite{costco2019} as $\Theta_{p}$ to predict the values of the augmented tensor cells, since \costco value predictor exhibits high prediction accuracy than other methods empirically (see Section~\ref{sec:exp:ablation_study} and Figure~\ref{fig:ablation_value}).

After we obtain all tensor cell indices and values needed for augmentation, we add them to the input tensor to obtain an augmented tensor $\tensor{X}_{new}$ (line 15).
This augmented tensor can be used for downstream tasks. 

\subsection{Complexity Analysis}
\label{sec:method:complexity}
In this subsection, we analyze the time and space complexity of \method. 

{\textbf{Time complexity.}} The first step of \method (Section~\ref{sec:NN_training}) is training a neural tensor completion model $\Theta$ to generate entity embeddings and gradients, and it takes $O(T_{\Theta})$ assuming $O(T_{\Theta})$ is the time complexity of training $\Theta$ as well as gradient calculations. 
    The second step of \method (Section~\ref{sec:instance_importance}) is computing the cell importance $\tensor{\alpha}_{z}$ for each training cell $z \in \Omega_{train}$. 
    A naive computation of $\tensor{\alpha}_{z}$ in Equation~\eqref{eq:instance:importance} for all training cells takes $O(KD|\Omega_{train}||\Omega_{val}|)$, where $K$ and $D$ are the number of checkpoints and the dimension of the gradient vector, respectively. We accelerate the computation to $O(KD(|\Omega_{train}|+|\Omega_{val}|))$ by precomputing $\sum_{z' \in \Omega_{val}} \nabla \ell(\Theta_{t_i},z')$ for all checkpoints $\Theta_{t_1}, ..., \Theta_{t_K}$ in Equation~\eqref{eq:instance:importance}. 
    The third step of \method (Section~\ref{sec:method:entity_importance}) is calculating the entity importance with the aggregation technique by Equation~\eqref{eq:entity_importance}. It takes $O(N|\Omega_{train}|)$ since we distribute $\tensor{\alpha}_{z}, \forall z \in \Omega_{train}$ to its entities and aggregate the assigned values. Finally, the data augmentation step (Section~\ref{sec:data_augmentation}) takes $O(T_{\Theta_{p}}+N_{aug}(N\log{I}+T_{infer}))$, where $O(T_{\Theta_{p}})$ and $O(T_{infer})$ are the training and single inference time complexities of the value prediction model $\Theta_{p}$, respectively. $O(N_{aug}N\log{I})$ term indicates the time complexity of weighted sampling $N_{aug}$ cells without replacement~\cite{wong1980efficient} from $N$ dimensions (assuming $I_1 = \cdots = I_N = I$). The final time complexity of \method is $O(T_{\Theta} + T_{\Theta_{p}} + (KD+N) |\Omega_{train}|+ KD|\Omega_{val}| + N_{aug}(N\log{I}+T_{infer}))$.
    
    {\textbf{Space complexity.}} The first step of \method (Section~\ref{sec:NN_training}), obtaining entity embeddings and gradients, takes $O(M_{\Theta}+KD(|\Omega_{train}|+|\Omega_{val}|))$ space, assuming $O(M_{\Theta})$ is the space complexity of training a neural tensor completion model $\Theta$ (including entity embeddings). $O(KD(|\Omega_{train}|+|\Omega_{val}|))$ space is required to store $D$-dimension gradients of training and validation cells for all $K$ checkpoints.
    The second step of \method (Section~\ref{sec:instance_importance}), computing the cell importance $\tensor{\alpha}_{z}$, takes  $O(|\Omega_{train}|)$ space since we need to store all cell importance values. The third step of \method (Section~\ref{sec:method:entity_importance}), calculating the entity importance with the aggregation method, takes $O(NI)$ space since we need to store importance scores of all entities from $N$ dimensions (assuming $I_1 = \cdots = I_N = I$). Finally, the data augmentation step (Section~\ref{sec:data_augmentation}) takes $O(M_{\Theta_{p}}+N_{aug}N)$ space since we need $O(M_{\Theta_{p}})$ space for training the value predictor $\Theta_{p}$ and $O(N_{aug}N)$ space for storing the data augmentation with $N_{aug}$ cells. The final space complexity of \method is $O(M_{\Theta} + M_{\Theta_{p}}+  KD(|\Omega_{train}|+ |\Omega_{val}|) + N(I+N_{aug}))$.
	
	\section{Experiments}
	\label{sec:experiment}
	In this section, we evaluate \method\ to answer the following questions. 
\begin{enumerate}
	\item{\textbf{Effectiveness of \method (Section~\ref{sec:exp:augmentation}).}
	How much does our proposed data augmentation technique enhance the accuracy of neural tensor completion compared to the baseline augmentation methods?
	}
	\item{\textbf{Ablation studies of \method (Section~\ref{sec:exp:ablation_study}).}
	\method consists of three major components: training entity embeddings, generating new tensor cells, and predicting values of the new cells.
	How much does each component of \method contribute to boosting the neural tensor completion accuracy? 
	}
	\item{\textbf{Comparisons of influence estimators (Section~\ref{sec:exp:comparison_inf}).}
	How much do different influence estimators impact the prediction accuracy improvements of \method?
	}
	\item{\textbf{Comparisons of entity importance algorithms (Section~\ref{sec:exp:comparisons_entity}).}
	How much do different entity importance methods affect the prediction accuracy improvements of \method?
	}
	\item{\textbf{Scalability of \method (Section~\ref{sec:exp:runtime}).}
    Does the running time of \method linearly scale with the number of data augmentation?
	}
	\item{\textbf{Hyperparameter sensitivity (Section~\ref{sec:exp:hyperparameter}).}
	How much do model hyperparameters of \method, such as embedding dimension and layer structures, affect the prediction accuracy?
	}
\end{enumerate}

We first describe the datasets and experimental settings in Section~\ref{sec:exp:settings}, and then answer the above questions. 

	\begin{table}[h!]
		\centering
		\caption{Summary of real-world tensor datasets used in our experiments.}
		\centering
		\begin{tabular}{| c | c | c | c | c |}
			\toprule
			\textbf{Name} & \textbf{Order} & \textbf{Dimensionality} & \textbf{Nonzeros} \\
			\midrule
			MovingMNIST & 4 & (100, 20, 64, 64)  & 819,200\\
			Foursquare & 3 & (2321, 5596, 378) & 388,210\\
			Reddit & 3 & (10000, 984, 744)  & 336,222\\
			LastFM & 3 & (980, 10000, 1587) &  258,620\\
		\bottomrule
		\end{tabular}	
		\label{tab:dataset}
	\end{table}

\subsection{Experimental Settings}
\label{sec:exp:settings}
\subsubsection{Datasets}
\label{sec:datasets}
We use four real-world tensor datasets to evaluate the performance of our proposed data augmentation method and baselines.
As summarized in Table~\ref{tab:dataset}, we use MovingMNIST~\cite{MovingMNIST}, Foursquare~\cite{yuan2013time}, Reddit~\cite{Reddit}, and LastFM~\cite{LastFM}.
MovingMNIST is a video tensor represented by (sequence, length, width, height; intensity), and we use 10\% of the total data.
Foursquare is a point-of-interest tensor represented by (user, location, timestamp; visit).
Reddit includes the posting history of users on subreddits represented by (user, subreddit, timestamp; posted). 
LastFM contains the music playing history of users represented by (user, music, timestamp; played).
For all tensors having timestamps, we converted their timestamps to unique days, so that all timestamps in the same day would have the same converted number. We randomly added negative samples with random indices and zero values to original data for Foursquare, Reddit, and LastFM tensors.

\subsubsection{Baselines}
\label{sec:competitors}
To the best of our knowledge, there are no existing methods for data augmentation designed specifically for tensors.
Therefore, we create a few baselines based on the broader literature in data augmentation by the following:
\begin{itemize}
	\item{\textbf{Duplication or Oversampling:}
		This simple data augmentation method randomly copies tensor cells and values from the existing training data, and adds them to the original tensor; there can be multiple tensor cells with the same values. 
	}
	\item{\textbf{Entity Replacement:}
		This data augmentation method randomly selects existing tensor cells and replaces their indices one by one with one of the top-10 closest entities in the embedding space, while keeping their tensor values.
	}
	\item{\textbf{(Random, MLP) and (Random, \costco):}
		These baselines generate data augmentation by randomly sampling missing tensor cells and predicting their values via the MLP and \costco~\cite{costco2019} models trained with the original tensor, respectively.
    }
\end{itemize}

Recall that neural tensor completion methods such as \costco~\cite{costco2019} and NTF~\cite{NTF2019} cannot generate new data points by themselves. Thus, these methods are not included in our baselines. The influence estimator \tracin~\cite{tracin} also only computes the cell importance but cannot create new data augmentation. We exclude non-neural tensor completion algorithms~\cite{ptucker,liu2014generalized,ShinK17,yang2021mtc} to test the data augmentation since \method is optimized for neural tensor completion methods.

\begin{figure}[t!]
	\centering
	\hspace{-3mm}
	\begin{subfigure}[t]{0.22\textwidth}
		\includegraphics[width=4.5cm]{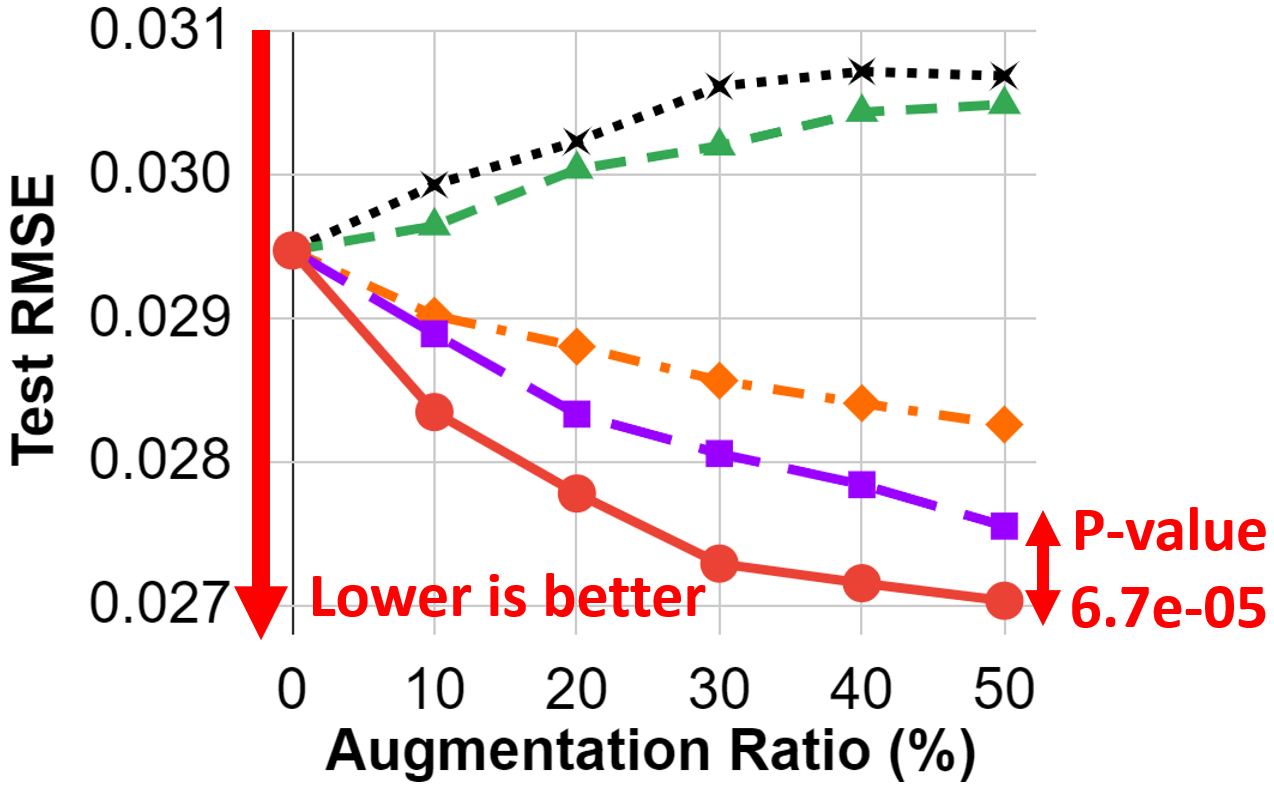}
		\captionsetup{justification=centering}
		\caption{MovingMNIST dataset}
		\label{fig:aug_mnist}
	\end{subfigure}
	\hspace{5mm}
	\begin{subfigure}[t]{0.22\textwidth}
		\includegraphics[width=4.5cm]{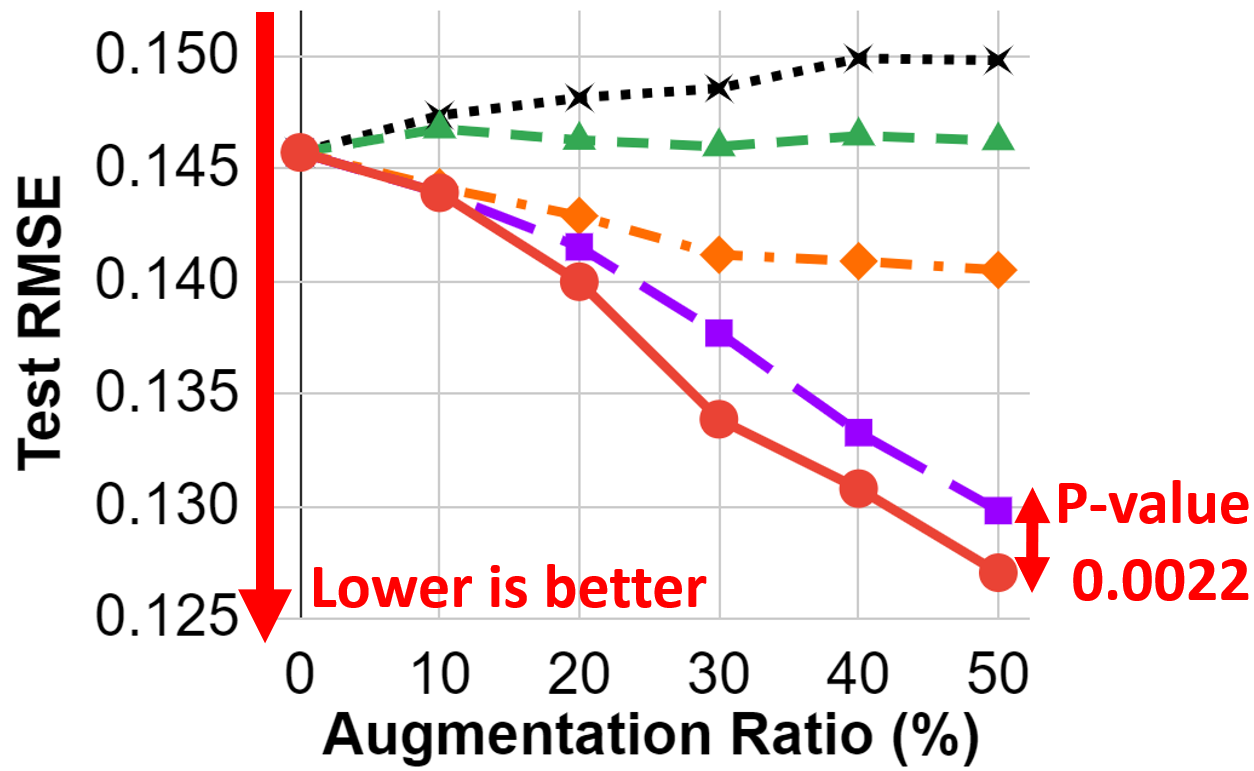}
		\captionsetup{justification=centering}
		\caption{Foursquare dataset}
		\label{fig:aug_foursquare}
	\end{subfigure}
	\\
	\hspace{-3mm}
	\begin{subfigure}[t]{0.22\textwidth}
		\includegraphics[width=4.5cm]{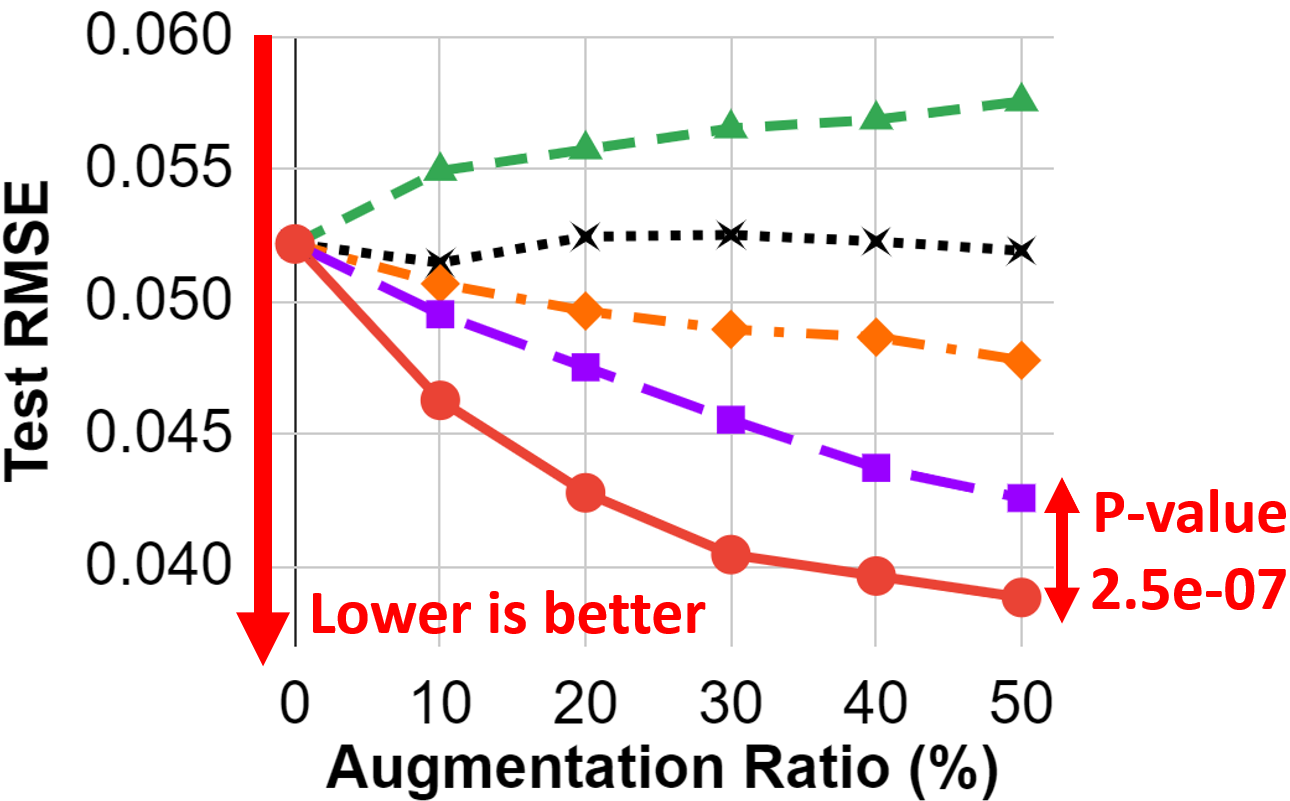}
		\captionsetup{justification=centering}
		\caption{Reddit dataset}
		\label{fig:aug_reddit}
	\end{subfigure}
	\hspace{5mm}
	\begin{subfigure}[t]{0.22\textwidth}
		\includegraphics[width=4.5cm]{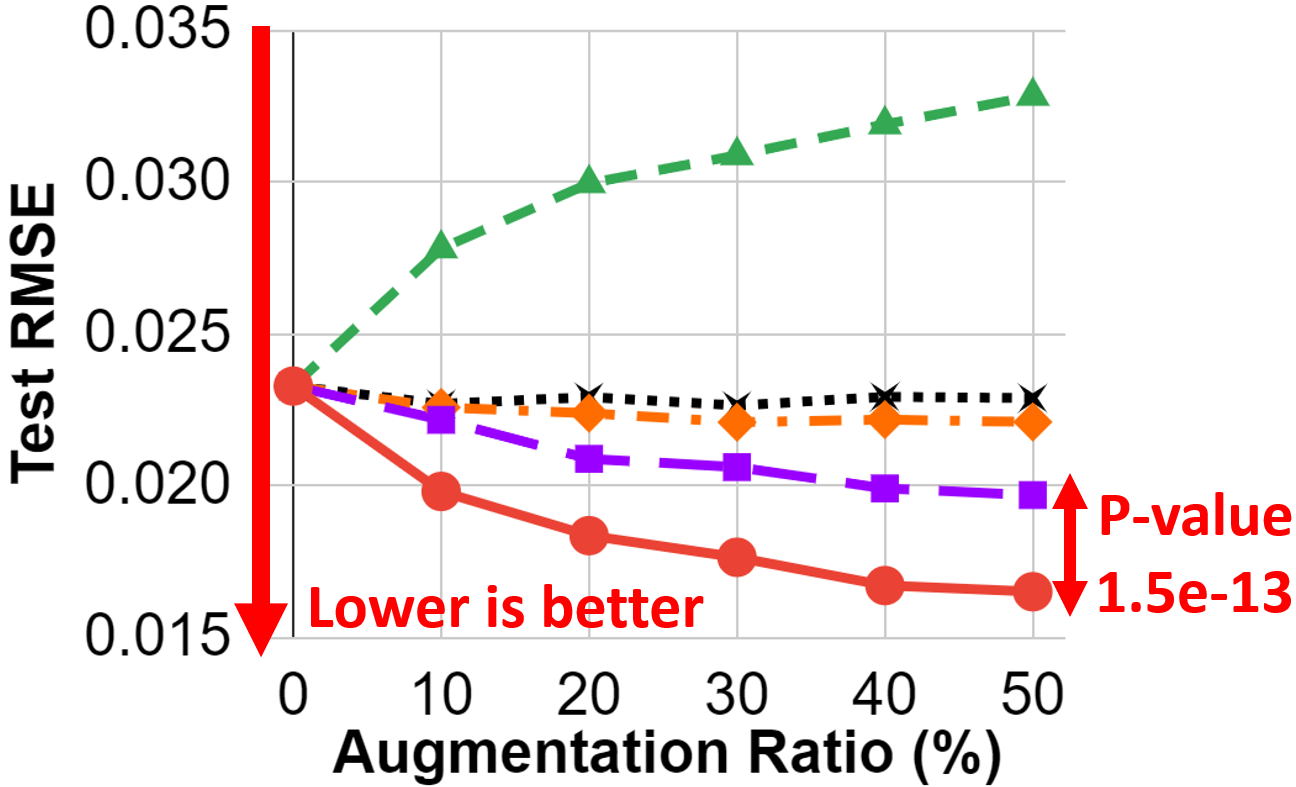}
		\captionsetup{justification=centering}
		\caption{LastFM dataset}
		\label{fig:aug_LastFM}
	\end{subfigure}
	\begin{subfigure}[t]{0.86\linewidth}
	\vspace*{2mm}
	\hspace*{-10mm}
	\includegraphics[width=9cm]{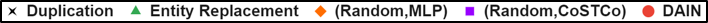}    
	\end{subfigure}
	\\
	\hspace{-3mm}
	\caption{\textit{Effectiveness of \method.} Test RMSE of the neural tensor completion method (MLP) after applying various data augmentation methods on real-world datasets. For all tensors, \method\ outperforms all baselines  and shows the lowest test RMSE value at the maximum augmentation with statistical significance.
	}
	\label{fig:augmentation}
\end{figure}

\subsubsection{Experiment Setup}
\label{sec:exp:setup}
We randomly split our datasets to use 90\% data for training and 10\% for test. We randomly sample 20\% of training data as the validation set. We stop the training of a neural network when the validation accuracy is not improving anymore, with a patience value of 10. 
We repeat all experiments 10 times and report the average test RMSE.
To measure statistical significance, we use a two-sided test of the null hypothesis that two independent samples have identical average (expected) values. If we observe a small p-value (\textit{e.g.}, $<0.01$), we can reject the null hypothesis. 

We fine-tune neural network hyperparameters of \method with validation data. Specifically, embedding dimension, batch size, layer structure, and learning rate are set to 50, 1024, [1024,1024,128], 0.001, respectively. 
We use Adaptive Moment Estimation (Adam)~\cite{adam} optimizer for the neural network training, and
the maximum number of epochs for training a neural network is set to 50. 
To compute the Cell Importance Tensor, we set checkpoints and step size to every epoch and 0.001, respectively. 
We take gradients with respect to the last fully connected layer.
We execute all our experiments on Azure Standard-NC24 machines equipped with 4 NVIDIA Tesla K80 GPUs with Intel Xeon E5-2690 v3 processor. 
The data augmentation module is implemented in Python with the PyTorch library.

\begin{figure}[t!]
	\centering
	\hspace{-6mm}
	\begin{subfigure}[t]{0.22\textwidth}
		\includegraphics[width=4cm]{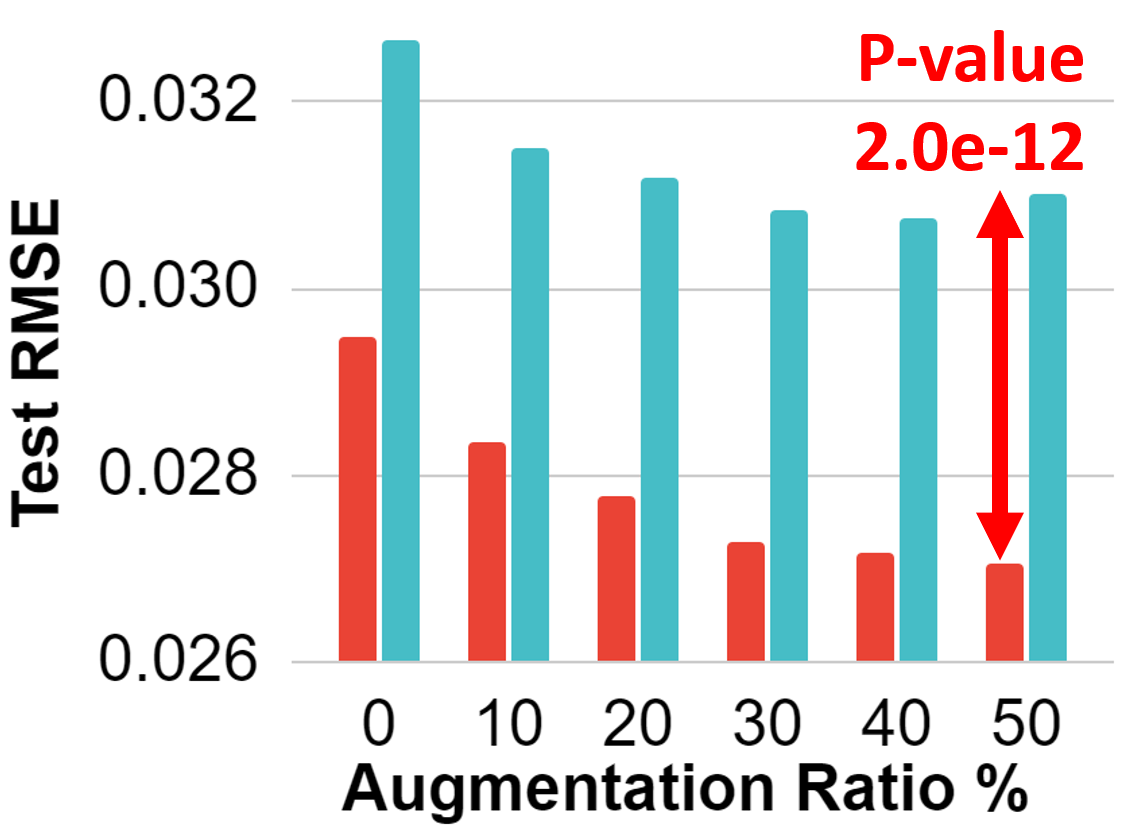}
		\captionsetup{justification=centering}
		\caption{MovingMNIST dataset}
		\label{fig:ablation_embedding_mnist}
	\end{subfigure}
	\hspace{5mm}
	\begin{subfigure}[t]{0.22\textwidth}
		\includegraphics[width=4cm]{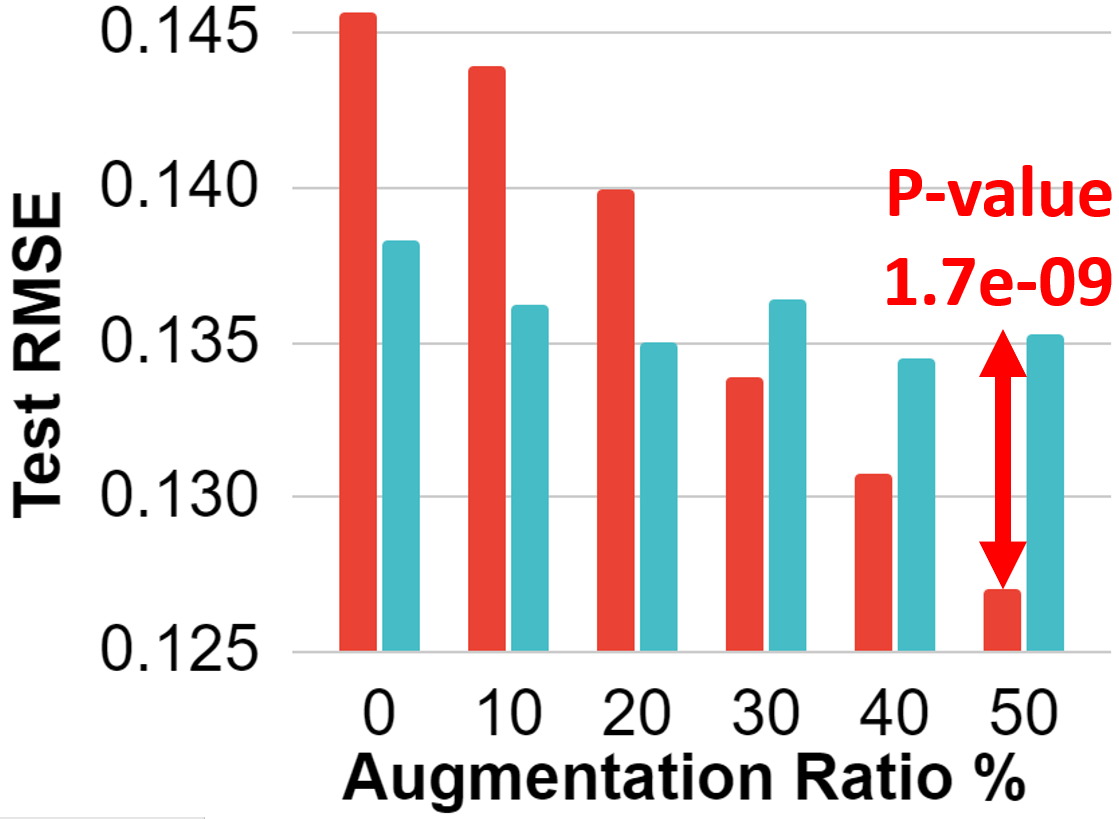}
		\captionsetup{justification=centering}
		\caption{Foursquare dataset}
		\label{fig:ablation_embedding_foursquare}
	\end{subfigure}
    \\
	\vspace{2mm}
	\includegraphics[width=8cm]{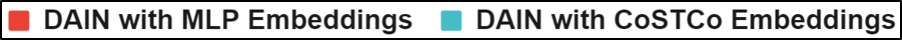}
	\caption{\textit{Ablation study of entity embedding generators of \method.} We test the effectiveness of the MLP embedding model of \method on MovingMNIST and Foursquare tensors. The MLP model  outperforms the \costco embedding model in terms of test RMSE at the maximum augmentation with statistical significance.
	}
	\label{fig:ablation_study_embedding}
\end{figure}

\begin{figure}[t!]
	\centering
	\hspace{-6mm}
	\begin{subfigure}[t]{0.22\textwidth}
		\includegraphics[width=4cm]{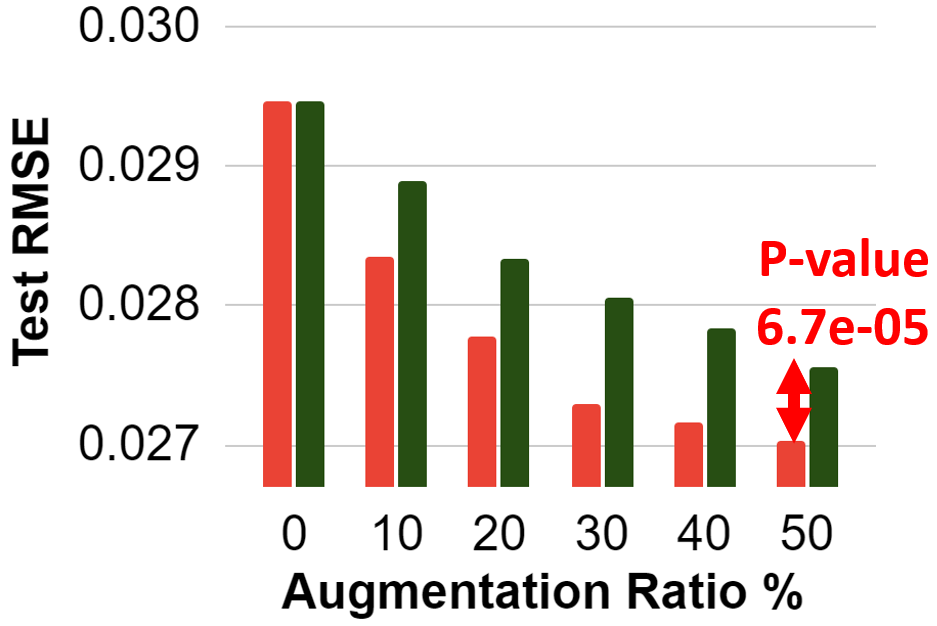}
		\captionsetup{justification=centering}
		\caption{MovingMNIST dataset}
		\label{fig:ablation_index_mnist}
	\end{subfigure}
	\hspace{5mm}
	\begin{subfigure}[t]{0.22\textwidth}
		\includegraphics[width=4cm]{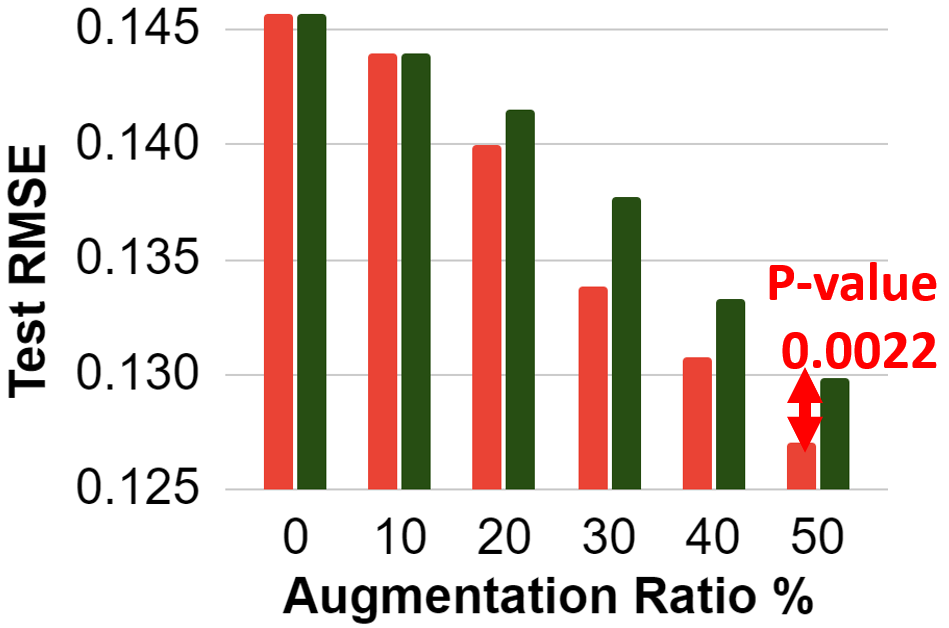}
		\captionsetup{justification=centering}
		\caption{Foursquare dataset}
		\label{fig:ablation_index_foursquare}
	\end{subfigure}
    \\
	\vspace{2mm}
	\includegraphics[width=9cm]{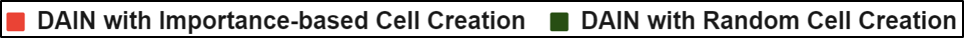}
	\caption{
	\textit{Ablation study of cell creation methods of \method.} We test the effectiveness of the cell creation component of \method on MovingMNIST and Foursquare tensors. The entity importance-based cell creation of \method outperforms the random cell creation baseline in terms of test RMSE at the maximum augmentation with statistical significance.
	}
	\label{fig:ablation_study_index}
\end{figure}

\begin{figure}[t!]
	\centering
	\hspace{-6mm}
	\begin{subfigure}[t]{0.22\textwidth}
		\includegraphics[width=4cm]{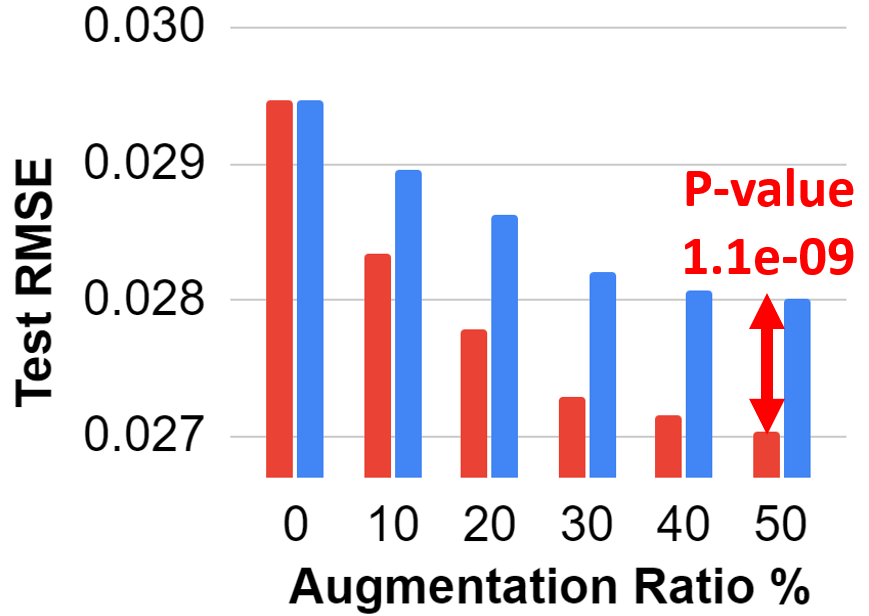}
		\captionsetup{justification=centering}
		\caption{MovingMNIST dataset}
		\label{fig:ablation_value_mnist}
	\end{subfigure}
	\hspace{5mm}
	\begin{subfigure}[t]{0.22\textwidth}
		\includegraphics[width=4cm]{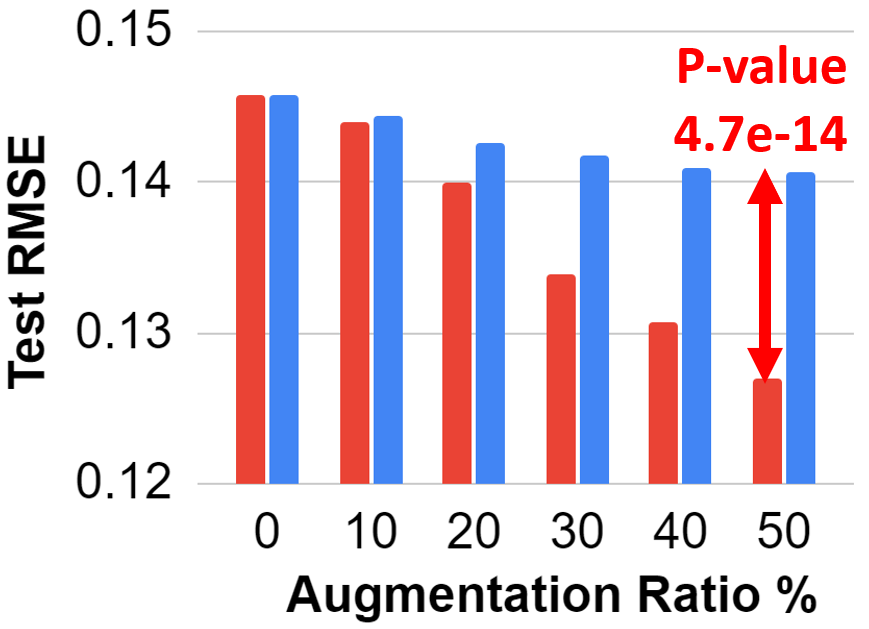}
		\captionsetup{justification=centering}
		\caption{Foursquare dataset}
		\label{fig:ablation_value_foursquare}
	\end{subfigure}
	\\
	\vspace{2mm}
	\includegraphics[width=7.5cm]{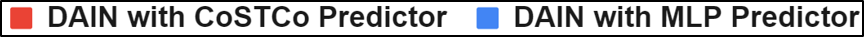}
	\caption{\textit{Ablation study of value prediction methods of \method.} We test the effectiveness of the value prediction component of \method on MovingMNIST and Foursquare tensors. The value prediction module of \method outperforms the MLP baseline in terms of test RMSE at the maximum augmentation with statistical significance. We choose the MLP baseline since it shows higher prediction accuracy than the random one.
	\label{fig:ablation_value}
	}
	\label{fig:ablation_study_value}
\end{figure}

\subsection{Effectiveness of \method}
\label{sec:exp:augmentation}
To test the effectiveness of our proposed data augmentation method, we measure and compare test RMSE values of the neural tensor completion model (MLP) after applying \method and baselines on real-world tensors.  
Figure~\ref{fig:augmentation} shows their performance. We omit error bars since they have small standard deviation values.

Both the baseline methods, Duplication and Entity Replacement, lead to a reduction in performance, \textit{i.e.}, an increase in test error, with any amount of augmentation. More augmentation leads to a greater reduction in performance. A potential reason is that the feature spaces of original and augmentation are very similar, so the augmentation cannot increase the generalization capability of a model. 
Although the other baselines (Random, MLP) and (Random, \costco) outperform Duplication or Entity Replacement, they have limitations in reducing test RMSE compared to \method.

Our proposed method \method leads to improvements in performance across all datasets with augmentation. 
\method performs the best across all five methods and presents the lowest test RMSE at the maximum augmentation with statistical significance, which clearly demonstrates the effectiveness of \method. The performance improves as more data is added to the original one. 
A key reason for the high performance of \method is that it combines important entities from each dimension, which is leading to the augmentation of more influential and heterogeneous data points to the original tensor. 

\subsection{Ablation Studies of \method}
\label{sec:exp:ablation_study}
At a high-level view, \method consists of three components---the first is training entity embeddings with a neural tensor completion method $\Theta$ (step 1 in Figure~\ref{fig:mainplot}), the second component is selecting new tensor cells for augmentation based on entity importance (steps 2 and 3), and the final is predicting values of the augmented cells using a neural tensor completion method $\Theta_{p}$ (step 4). 
As mentioned previously, in \method, we use MLP, entity importance, and \costco-based predictor in the three components, respectively. \footnote{Please note that we also conducted experiments with \costco as the embedding generator and MLP as the augmented cell's value predictor, but its prediction performance was significantly lower (with $p$-value < 0.05). Thus, in this subsection, we conduct ablation studies on the proposed model where MLP is the embedding generator and \costco is the value predictor.}
We perform ablation studies to investigate the contribution of each component in \method's performance by replacing the component with the baseline component, while fixing all the others. 
Figures~\ref{fig:ablation_study_embedding}, \ref{fig:ablation_study_index}, and \ref{fig:ablation_study_value} show ablation study results of the three components of \method on the two largest tensor datasets, namely MovingMNIST and Foursquare.

In Figure~\ref{fig:ablation_study_embedding}, \method with MLP embedding generator presents lower test RMSE than a baseline of \method with \costco embedding generator at 50\% augmentation ratio. A potential reason is that using \costco in both step 1 and step 4 (for augmented cell's value prediction) can lead to overfitting. On the other hand, using MLP in step 1 and \costco in step 4 gives generalization capability to the model. The entity embedding module has a huge impact on performance improvements since entity embeddings and gradients generated by the module affect all the subsequent modules.

In Figure~\ref{fig:ablation_study_index}, \method with entity importance-based cell creation beats the model with random cell creation. The result substantiates the usefulness of using entity importance in creating tensor cells for augmentation. 

In Figure~\ref{fig:ablation_study_value}, \method with \costco for augmented cell's value prediction outperforms \method with a MLP-based value predictor. This is again because of overfitting where the MLP predictor produces homogeneous data points since the entity embedding model in step 1 is an MLP as well. On the other hand, using a \costco-based value predictor produces heterogeneous data points giving generalizability. 

In summary, all of the components of \method prove useful by reducing test RMSE values significantly compared to baseline components. 

\begin{figure}[t!]
	\centering
	\hspace{-3mm}
	\begin{subfigure}[t]{0.22\textwidth}
		\includegraphics[width=4cm]{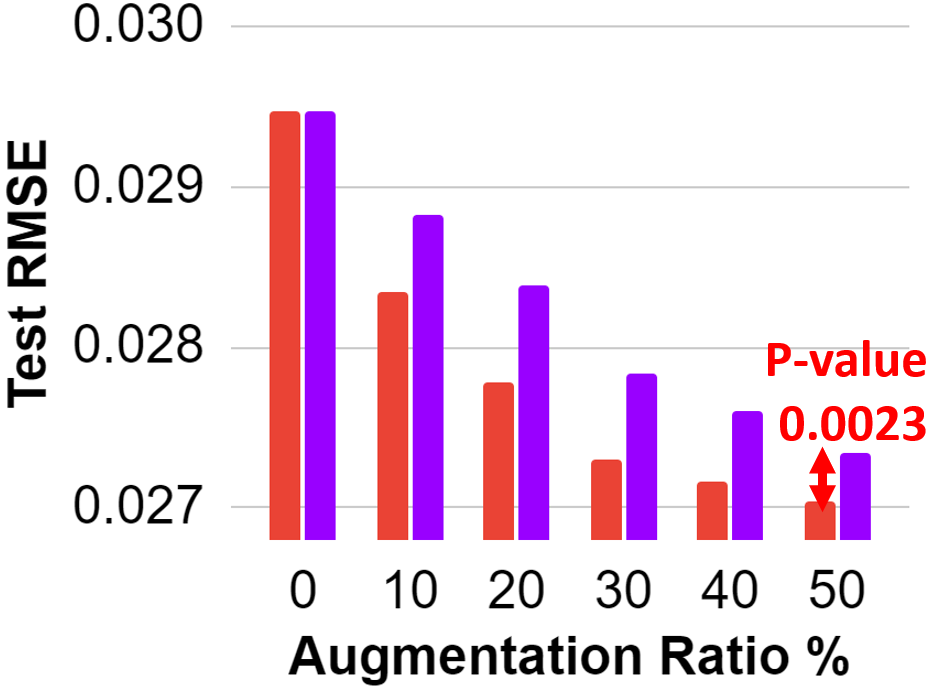}
		\captionsetup{justification=centering}
		\caption{MovingMNIST dataset}
		\label{fig:compare_inf_mnist}
	\end{subfigure}
	\hspace{5mm}
	\begin{subfigure}[t]{0.22\textwidth}
		\includegraphics[width=4cm]{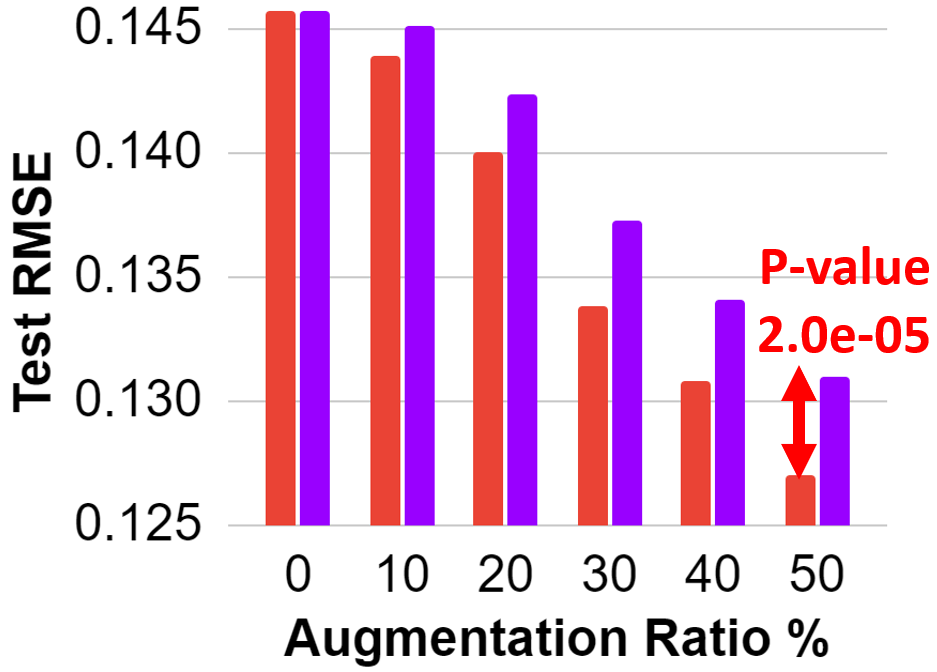}
		\captionsetup{justification=centering}
		\caption{Foursquare dataset}
		\label{fig:compare_inf_foursquare}
	\end{subfigure}
	\\
	\vspace{2mm}
	\includegraphics[width=6.5cm]{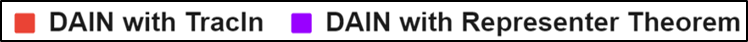}
	\caption{\textit{Comparisons of influence estimators.} We compare two influence estimators with respect to test RMSE improvements on MovingMNIST and Foursquare tensors. We find the \tracin influence estimator is more suitable for our data augmentation framework.
	}
	\label{fig:exp:comparisons_influence_functions}
\end{figure}

\begin{figure}[t!]
 	\centering
	\hspace{-6mm}
	\begin{subfigure}[t]{0.22\textwidth}
		\includegraphics[width=4cm]{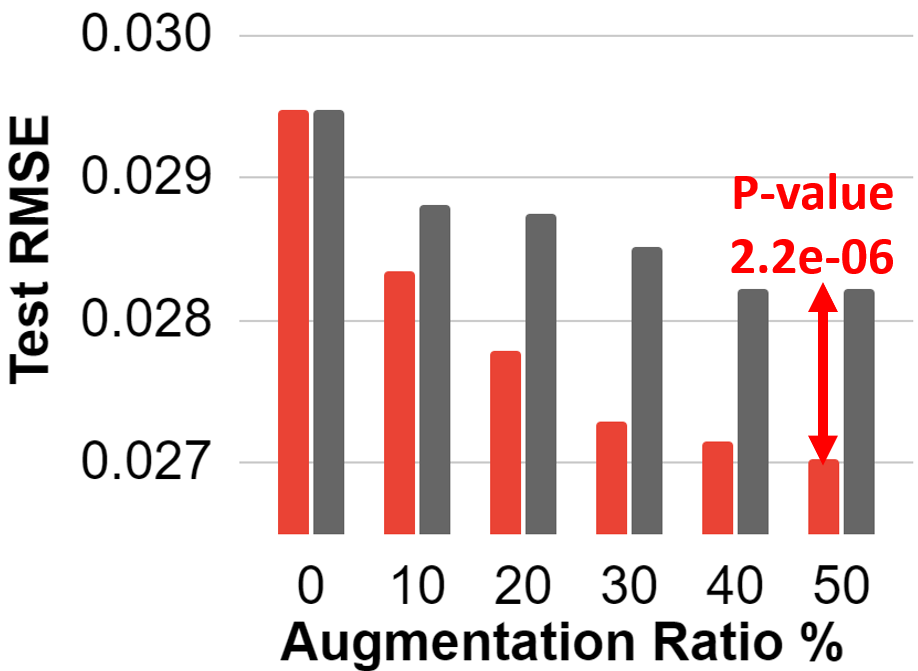}
		\captionsetup{justification=centering}
		\caption{MovingMNIST dataset}
		\label{fig:compare_entity_mnist}
	\end{subfigure}
	\hspace{5mm}
	\begin{subfigure}[t]{0.22\textwidth}
		\includegraphics[width=4cm]{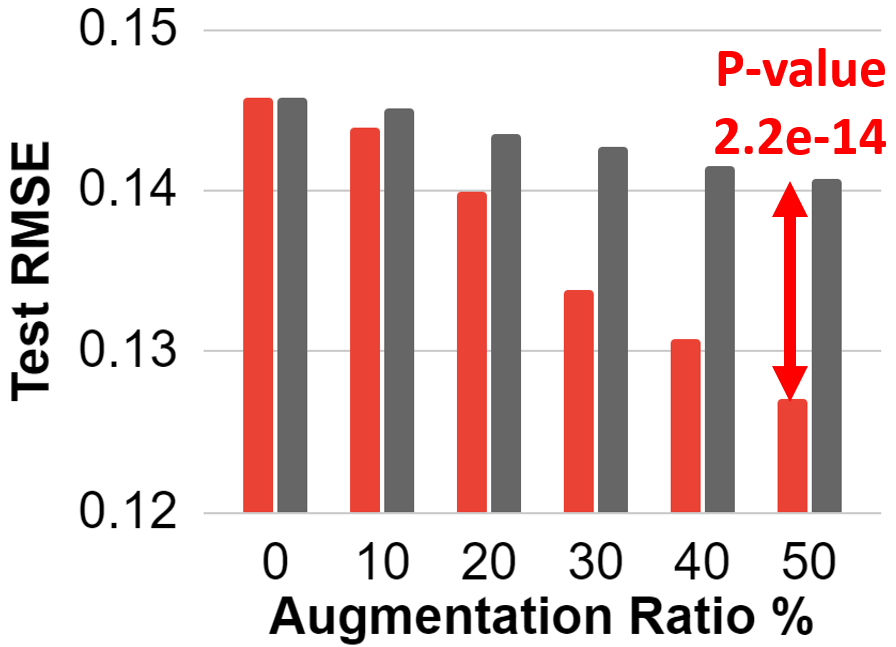}
		\captionsetup{justification=centering}
		\caption{Foursquare dataset}
		\label{fig:compare_entity_foursquare}
	\end{subfigure}
	\\
	\vspace{2mm}
	\includegraphics[width=7cm]{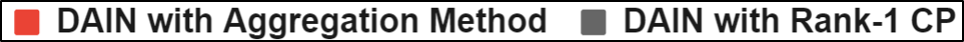}
	\caption{\textit{Comparisons of entity importance algorithms.} We compare two entity importance algorithms with respect to test RMSE improvements on MovingMNIST and Foursquare tensors. We find entity importance calculation by the aggregation scheme is more suitable for our data augmentation framework.
	}
	\vspace{-2mm}
	\label{fig:exp:comparisons_entity_importance}
\end{figure}

\subsection{Comparison of Influence Estimators}
\label{sec:exp:comparison_inf}
In this subsection, we explore how different influence estimators affect the prediction accuracy improvements of \method  on the two largest real-world tensors, namely MovingMNIST and Foursquare datasets.  
We compare two influence calculation methods: \tracin~\cite{tracin} and Representer Theorem~\cite{yeh2018representer}. We exclude influence function~\cite{KohL17} since it shows worse prediction accuracy than the Representer Theorem method. As shown in Figures~\ref{fig:compare_inf_mnist} and \ref{fig:compare_inf_foursquare}, \tracin consistently outperforms Representer Theorem with any amount of augmentation (with statistical significance at 50\% augmentation). One possible explanation for this gap is that \tracin directly computes the influence of a training cell in reducing validation loss, while Representer Theorem cannot measure the direct contribution.

\subsection{Comparison of Entity Importance Methods}
\label{sec:exp:comparisons_entity}
Similar to Section~\ref{sec:exp:comparison_inf}, we confirm how different entity importance calculators affect the prediction accuracy improvements of \method  on the two largest real-world tensors, namely MovingMNIST and Foursquare datasets.  
Between two entity importance calculators introduced in Section~\ref{sec:method:entity_importance}, the aggregation algorithm shows superior performance than the rank-1 CP factorization (see Figures~\ref{fig:compare_entity_mnist} and \ref{fig:compare_entity_foursquare}) with extensive test RMSE differences at 50\% augmentation. The main reason for this gap is that rank-1 CP factorization cannot decompose the Cell Importance Tensor accurately due to its high sparsity, while the aggregation algorithm does not suffer from the sparsity issue.

\subsection{Scalability of \method}
\label{sec:exp:runtime}
In this subsection, we measure the runtime of \method\ on the largest real-world, namely MovingMNIST, and a synthetic tensor, namely Synthetic-1M. We construct the synthetic tensor with dimensionality $(1000, 1000, 1000, 1000)$ and $1,000,000$ non-zero cells. We randomly generated factor matrices and reconstructed a synthetic tensor from them. 
Note that the total running time for $0\%$ augmentation is equivalent to the neural network training time. 
On the other hand, our proposed data augmentation method involves four runtime-related components: (1) initial neural network training, (2) cell importance calculation, (3) entity importance calculation, and (4) weighted sampling and value inference. Therefore, there are static costs (\textit{i.e.}, $O(T_{\Theta} + T_{\Theta_{p}} + (KD+N) |\Omega_{train}|+ KD|\Omega_{val}|$ in Section~\ref{sec:method:complexity}) resulted from steps (1), (2), and (3), and linearly increasing costs (\textit{i.e.}, $O(N_{aug}(N\log{I}+T_{infer}))$ in Section~\ref{sec:method:complexity}) proportional to the amount of augmentation for step (4). 

\begin{figure}[t!]
	\centering
	\hspace{-6mm}
	\begin{subfigure}[t]{0.23\textwidth}
		\includegraphics[width=4cm]{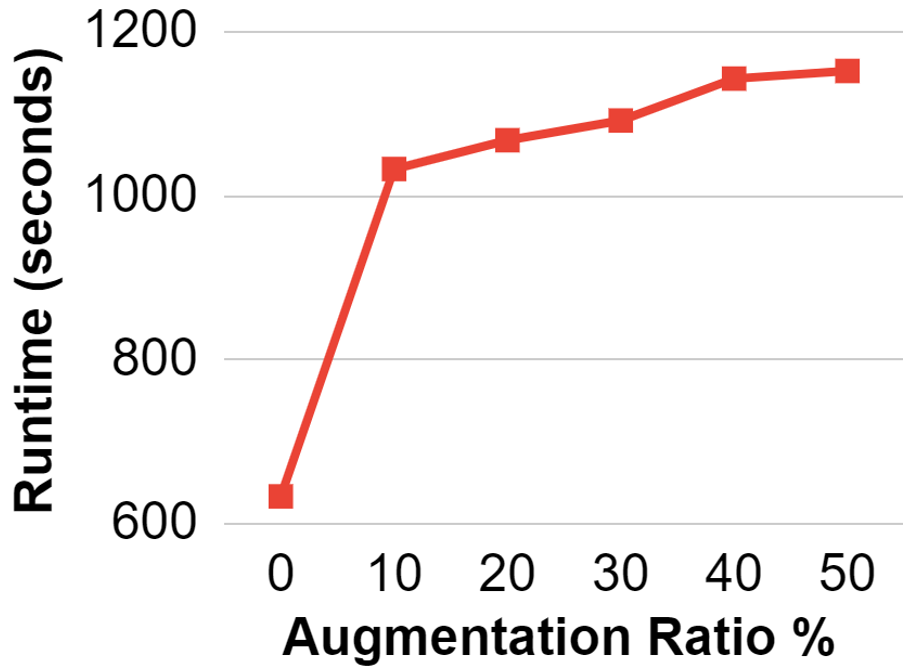}
		\captionsetup{justification=centering}
		\caption{MovingMNIST dataset}
		\label{fig:running_time_mnist}
	\end{subfigure}
	\hspace{3mm}
	\begin{subfigure}[t]{0.23\textwidth}
		\includegraphics[width=4cm]{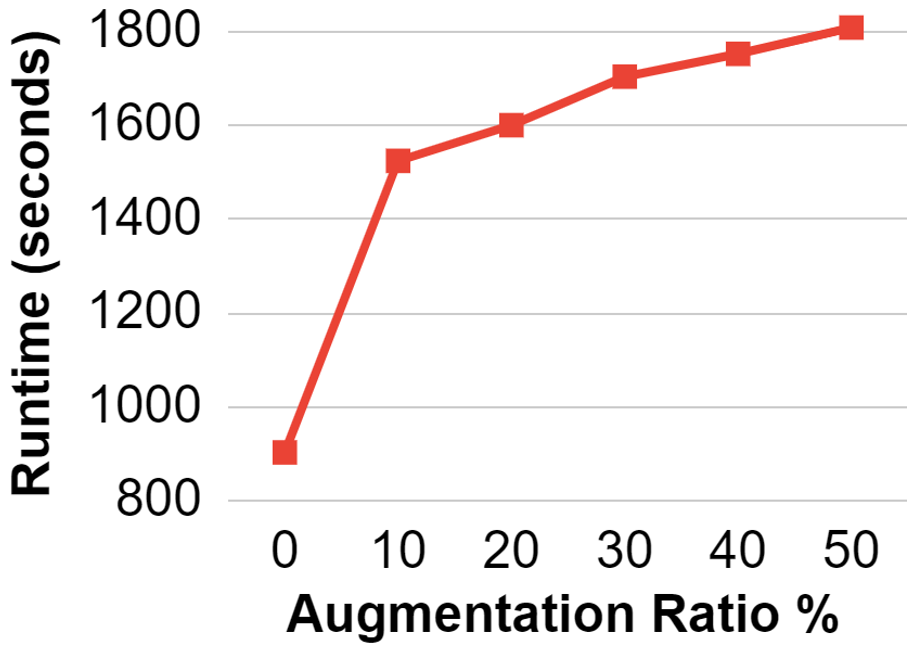}
		\captionsetup{justification=centering}
		\caption{Synthetic-1M dataset}
		\label{fig:running_time_synthetic}
	\end{subfigure}
	
	\caption{\textit{Scalability of \method.} This plot measures the running time of \method\ at different augmentation levels. The sudden jump in total running time from 0\% to 10\% augmentation occurs due to the cost for the cell and entity importance calculation (see time complexity analysis in Section~\ref{sec:method:complexity}); the runtime increases linearly thereafter. 
 	}
	\label{fig:runtime}
\end{figure}

Figure~\ref{fig:runtime} exhibits the total running time of \method\ on MovingMNIST and Synthetic-1M tensors. As expected before, we find a sudden increase in total running time from 0\% to 10\% augmentation on both tensors due to steps (2) and (3) above. After that, the running time increases linearly. This shows that \method\ is highly scalable and can be applied to large datasets.

\begin{figure}[t!]
	\centering
	\hspace{-6mm}
	\begin{subfigure}[t]{0.23\textwidth}
		\includegraphics[width=4cm]{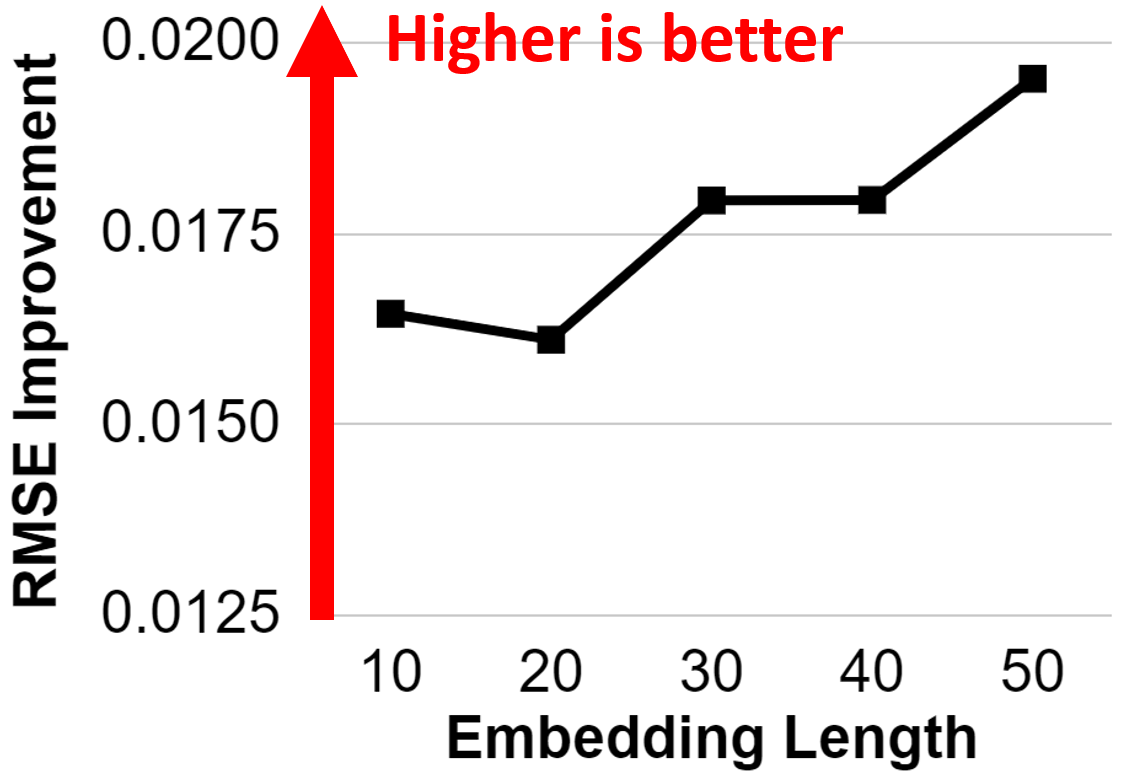}
		\captionsetup{justification=centering}
		\label{fig:hyper_emb_dim}
	\end{subfigure}
	\hspace{3mm}
	\begin{subfigure}[t]{0.23\textwidth}
		\includegraphics[width=4cm]{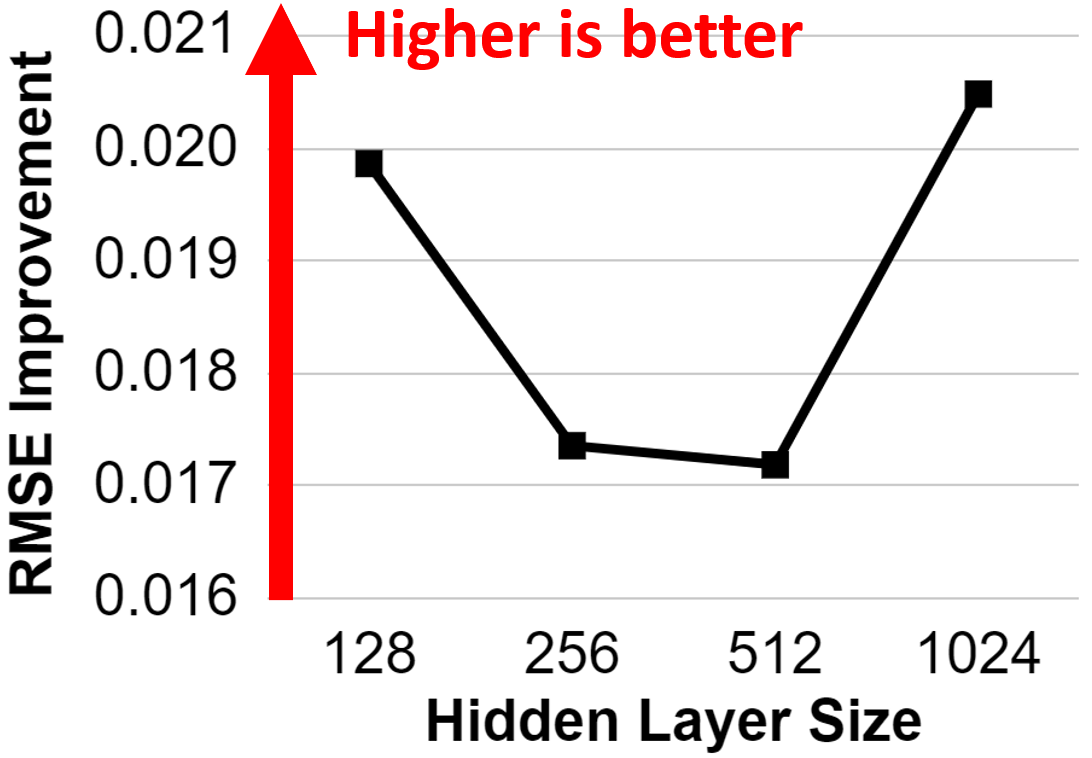}
		\captionsetup{justification=centering}
		\label{fig:hyper_layer_size}
	\end{subfigure}
	\\
	\vspace{1mm}
	\hspace{-6mm}
	\begin{subfigure}[t]{0.23\textwidth}
		\includegraphics[width=4cm]{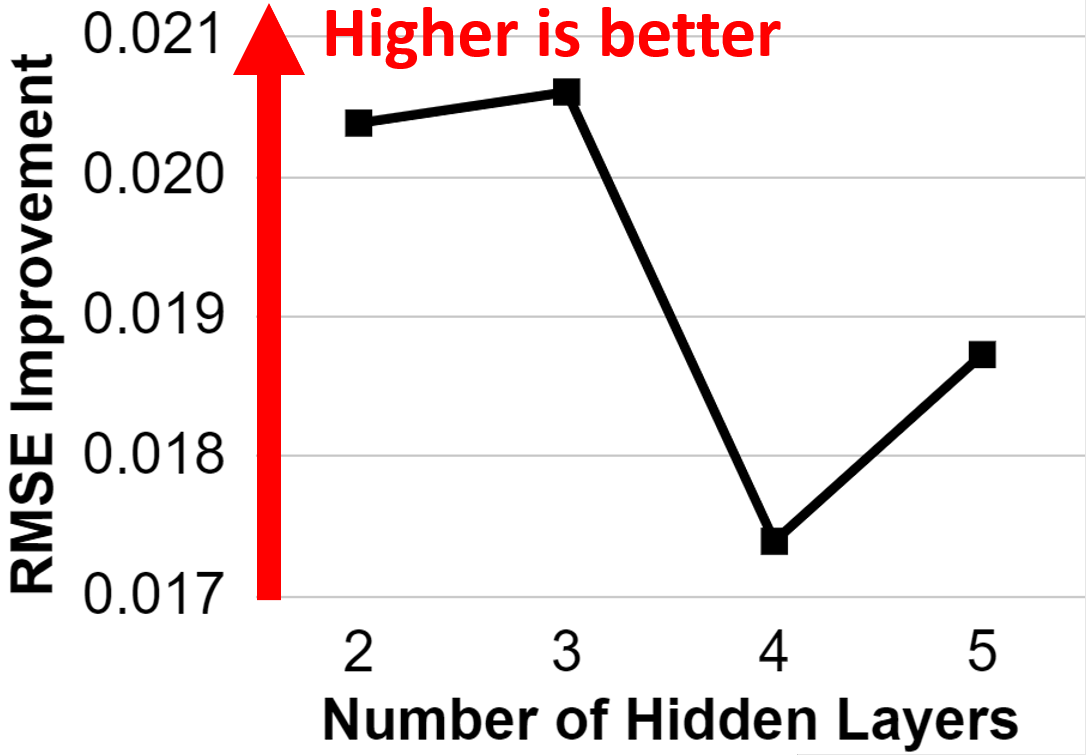}
		\captionsetup{justification=centering}
		\label{fig:hyper_num_of_layer}
	\end{subfigure}
	\hspace{3mm}
	\begin{subfigure}[t]{0.23\textwidth}
		\includegraphics[width=4cm]{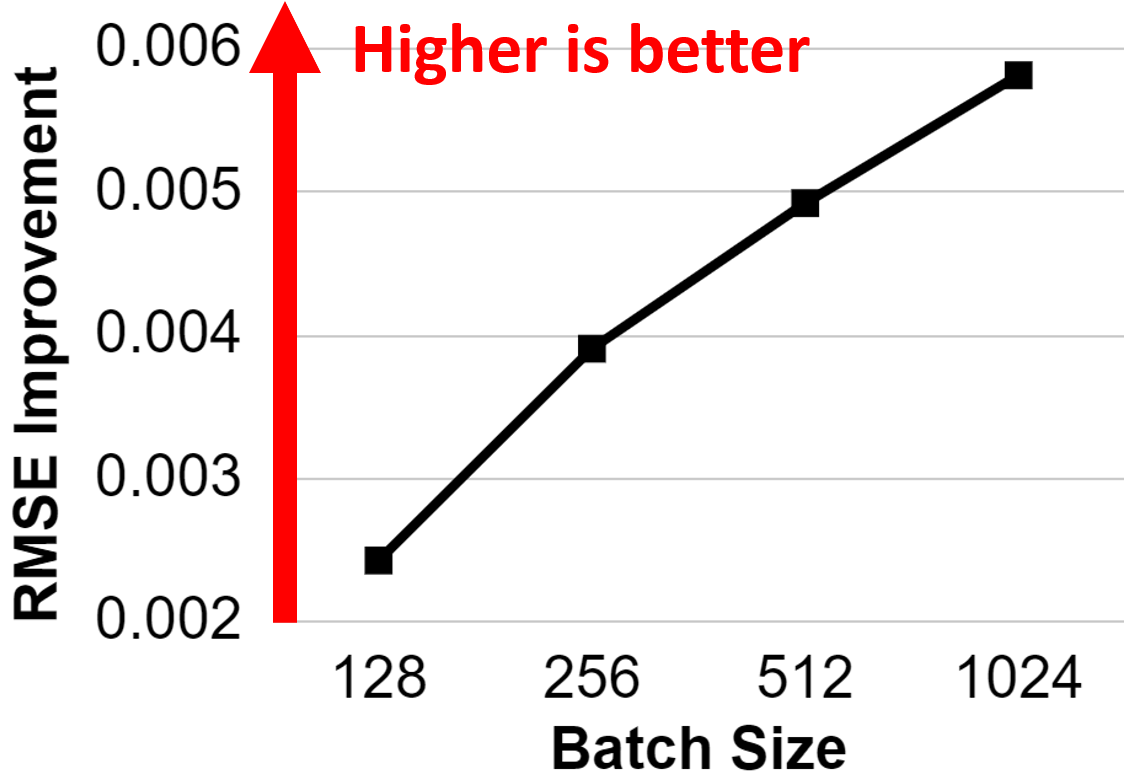}
		\captionsetup{justification=centering}
		\label{fig:hyper_batch_size}
	\end{subfigure}
	\\
	\vspace{1mm}
	\begin{subfigure}[t]{0.23\textwidth}
		\includegraphics[width=4cm]{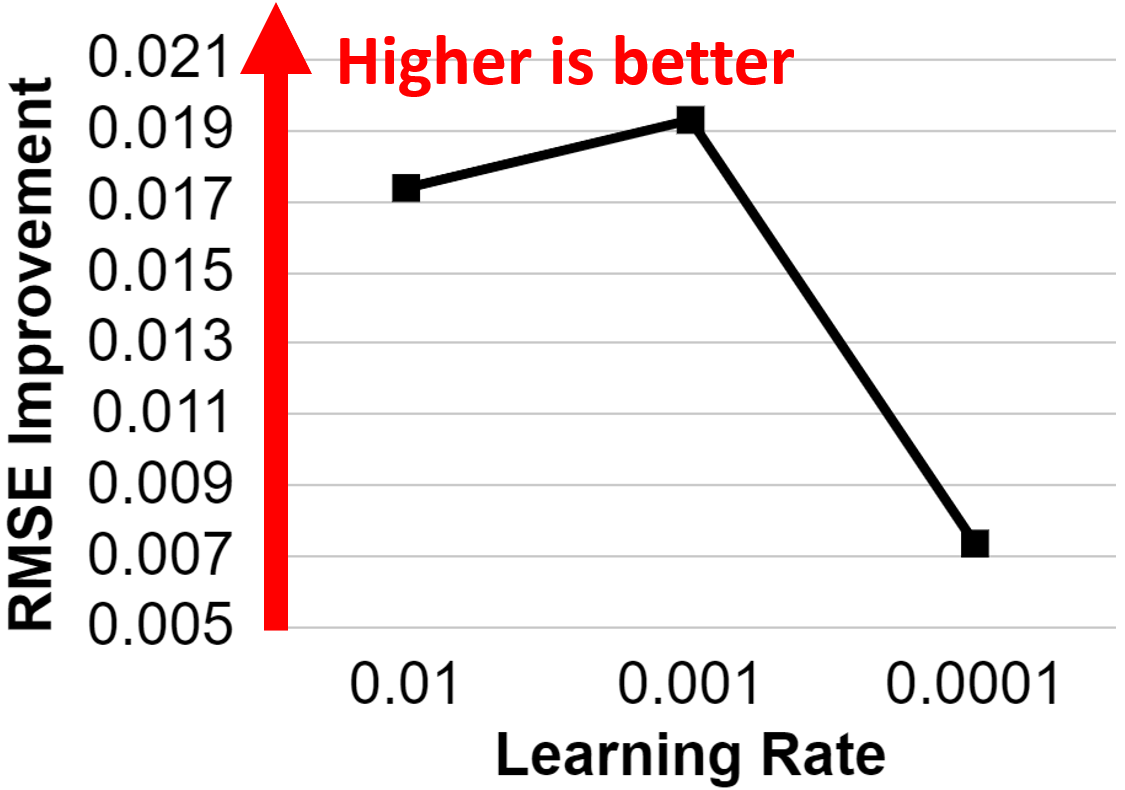}
		\captionsetup{justification=centering}
		\label{fig:hyper_lrate}
	\end{subfigure}
	\caption{\textit{Hyperparameter sensitivity plots.} We vary one of the hyperparameters of \method, namely embedding length, hidden layer size, number of hidden layers, batch size, and learning rate, while fixing all the others to default values. We measure the validation RMSE improvements after 50\% augmentation on the Foursquare dataset. We find medium-sized layers with large embedding dimensions, and proper learning rates are key factors for the prediction accuracy.
	}
	\label{fig:hyperparameter}
\end{figure}

\subsection{Hyperparameter Sensitivity}
\label{sec:exp:hyperparameter}
In this subsection, we investigate how much model hyperparameters affect the prediction accuracy of a model. Our neural network hyperparameters include the length of entity embeddings, the number of hidden layers, the size of a hidden layer, learning rate, and batch size.
We vary one hyperparameter while fixing all the others to default values mentioned in Section~\ref{sec:exp:setup}.

Figure~\ref{fig:hyperparameter} shows the hyperparameter sensitivity of \method with respect to the prediction accuracy on the Foursquare dataset with 50\% augmentation ratio. The RMSE improvement indicates how much validation RMSE values are enhanced after the augmentation (higher is better).
As the embedding size increases, we observe performance improvements since large embeddings contain more useful information about training data. 
Regarding the number of hidden layers, 3 hidden layers are appropriate since 1 or 2 layers may not fully learn the augmented tensor, and 4 layers may overfit. Too small or too large of learning rates degrade the prediction accuracy since a small one leads to slower convergence, and a large one can find a low-quality local optimum. 

	\section{Conclusion}
	\label{sec:conclusion}
	In this paper, we proposed a novel data augmentation framework \method\ for enhancing neural tensor completion.
The key idea is to augment the original tensor with new data points that are crucial in reducing the validation loss. 
Experimental results on real-world datasets show that \method\ outperforms baseline methods in various augmentation settings with statistical significance. Ablation studies of \method demonstrate the effectiveness of the key components of \method.  We also verify that \method\ scales near linearly to large datasets. 
Future directions of this work include exploring the effectiveness of \method\ in downstream tasks such as anomaly detection and dataset cleanup. Deriving theoretical guarantees of the performance boost from \method is worth exploring as well.

	\section*{Acknowledgment}
	{
	This research is supported in part by Adobe, Facebook, NSF IIS-2027689, Georgia Institute of Technology, IDEaS, and Microsoft. S.O. was partly supported by ML@GT, Twitch, and Kwanjeong fellowships. We thank the reviewers for their feedback.
	}	
	\bibliographystyle{ACM-Reference-Format}
	\bibliography{paper-DAIN}
	
\end{document}